\documentclass{article}

\usepackage{PRIMEarxiv}

\usepackage[utf8]{inputenc} 
\usepackage[T1]{fontenc}    
\usepackage{hyperref}       
\usepackage{url}            
\usepackage{booktabs}       
\usepackage{amsfonts}       
\usepackage{nicefrac}       
\usepackage{microtype}      
\usepackage{lipsum}
\usepackage{fancyhdr}       
\usepackage{graphicx}       
\graphicspath{{media/}}     

\usepackage{amssymb}
\usepackage{graphicx}
\usepackage{booktabs}
\usepackage{multirow}
 
\usepackage{amsmath}
\usepackage{caption}
\usepackage{subcaption}
\usepackage{algorithm}
\usepackage{algpseudocode}

\usepackage{amsmath,amssymb,amsfonts}
\usepackage{algorithm}
\usepackage{algpseudocode}
\usepackage{graphicx}
\usepackage{textcomp}
\usepackage{multirow}
\usepackage{longtable}
\usepackage{rotating} 
\usepackage{longtable}
\usepackage{booktabs}
\usepackage{ltxtable} 
\usepackage{supertabular,booktabs}
\usepackage[export]{adjustbox}
\usepackage{tablefootnote}
\usepackage{color}
\usepackage{xcolor}

\usepackage{changes}

\title{Privacy-Preserving Customer Churn Prediction Model in the Context of Telecommunication Industry
}

\author{
   Joydeb Kumar Sana \\
  Department of Computer Science and Engineering \\
  Bangladesh University of Engineering and Technology \\
  Dhaka, Bangladesh\\
  \texttt{joysana@gmail.com} \\
   \And
  M Sohel Rahman \\
  Department of Computer Science and Engineering \\
  Bangladesh University of Engineering and Technology \\
  Dhaka, Bangladesh\\
  \texttt{sohel.kcl@gmail.com} \\
   \And
  M Saifur Rahman \thanks{Corresponding author: mrahman@cse.buet.ac.bd} \\
  Department of Computer Science and Engineering \\
  Bangladesh University of Engineering and Technology \\
  Dhaka, Bangladesh\\
  \texttt{mrahman@cse.buet.ac.bd} \\
}

\begin{document}
\maketitle

\begin{abstract}
Data is the main fuel of a successful machine learning model. A dataset may contain sensitive individual records e.g. personal health records, financial data, industrial information, etc. Training a model using this sensitive data has become a new privacy concern when someone uses third-party cloud computing. Trained models also suffer privacy attacks which leads to the leaking of sensitive information of the training data. This study is conducted to preserve the privacy of training data in the context of customer churn prediction modeling for the telecommunications industry (TCI). In this work, we propose a framework for privacy-preserving customer churn prediction (PPCCP) model in the cloud environment. We have proposed a novel approach which is a combination of Generative Adversarial Networks (GANs) and adaptive Weight-of-Evidence (aWOE). Synthetic data is generated from GANs, and aWOE is applied on the synthetic training dataset before feeding the data to the classification algorithms. Our experiments were carried out using eight different machine learning (ML) classifiers on three openly accessible datasets from the telecommunication sector. We then evaluated the performance using six commonly employed evaluation metrics. In addition to presenting a data privacy analysis, we also performed a statistical significance test. The training and prediction processes achieve data privacy and the prediction classifiers achieve high prediction performance (87.1\% in terms of F-Measure for GANs-aWOE based Na\"{\i}ve Bayes model). In contrast to earlier studies, our suggested approach demonstrates a prediction enhancement of up to 28.9\% and 27.9\% in terms of accuracy and F-measure, respectively.
\end{abstract}


\section{Introduction}
Research in machine learning for decision support has gained significant attention as well as achievement in recent times, due to the availability of large amount of data generated from various sources.  Usually, these data are privacy sensitive both from the legal and ethical contexts. These data can be used to train various machine learning models to facilitate decision-making across various domains, encompassing healthcare, manufacturing, education, financial modeling, marketing, and more. For training a model, data processing, analyzing, and performing complex calculations on this large amount of data require huge memory, storage, computational resources as well as efficient algorithms. The cloud service is the cheapest and most available way to store and compute a vast amount of data. 

Over the past few years, the telecommunications industry (TCI) has experienced significant expansion and technological advancements. TCI industry generates a huge amount of customer data every day and they store the user data in their customer relationship management (CRM) system. The customer data can be used to train the churn prediction model. Due to intense competition, saturated markets, dynamic landscape, and attractive and profitable incentives from competitors, every player in the TCI is confronted with significant challenges related to customer churn, which is widely acknowledged as a formidable problem within this context \cite{b01}. Within this competitive market, customers have the flexibility to swiftly change services and even shift from one service provider to another. These customers, who avail these options, are commonly labeled as churned customers ~\cite{b01} in relation to their initial service provider. If a telecommunications company (TELCO) gains the ability to anticipate a customer's likelihood of churning, it can potentially cater specific offerings to that customer. This personalized offering approach aims to diminish dissatisfaction, enhance engagement, and consequently improve the chances of retaining the customer. Such actions would significantly contribute to boosting revenue, especially considering the relatively high costs associated with acquiring new customers ~\cite{b7, b34}. At present, TELCOs mainly prioritize retaining their existing long-term customers rather than acquiring new ones ~\cite{b12_Amin_Anwar}. To tackle the customer churn prediction (CCP) problem, several approaches have been presented in the literature (e.g.,  \cite{b12_Amin_Anwar},\cite{b13_Amin_Shehzad}, \cite{b14_Kirui}, \cite{Renjith_b19},   \cite{b15_Pendharkar},\cite{b16}, etc.). All the proposed CCP techniques are based on CRM data which is very privacy sensitive. To focus on their own business, data owners like TELCO often leverage third-party cloud computing facilities for data analysis and machine learning model development \cite{Ramzan2018}. Third-party computation during the model training using these raw data creates a privacy concern. For example, privacy can be breached due to dishonest or curious service providers. Data can also be leaked unintentionally due to data diversity and features during computation time. Sometimes the residual representation of the data is responsible for the data leakage. Adversaries can maliciously achieve unpremeditated but useful and sensitive data from the trained machine learning models via membership inference attacks, model inversion attacks, etc. \cite{Zhao2019, Papernot2018}. Sometimes personal information can be leaked due to the model over-fitting \cite{Yeom2018}. The sensitive private records should not be leaked from the trained model or during the model training. Protecting the privacy of data is difficult when data owners outsource the machine learning task to a cloud service provider as follows. First, training using plain text data is a security threat. Second, there is a trade-off between privacy and utility of the data, if we use differential privacy or add some noise to the dataset. Third, the cloud server can not work well if we use an encrypted dataset. Our main objective is to preserve data privacy while performing third-party computation without sacrificing performance.

Many techniques have been used to protect the sensitive user data from attackers and service providers. Usually, In order to preserve the privacy of the data, the data owner encrypts the sensitive data before uploading it to the cloud server. Subsequently, the cloud server performs some computations on the encrypted data. Public key- private key based encryption is common in the literature to preserve privacy of user data in cloud computation. The most trusted and common encryption schemes are based on homomorphic properties, which allow for certain mathematical operations on encrypted data without decryption.  For example, additive homomorphic property based Paillier cryptosystem \cite{Paillier_1999}  is used in  \cite{Erkin_2009}  for privacy-preserving face recognition. The BGN `doubly homomorphic' encryption \cite{Boneh_2005} is applied in \cite{Yuan_2014} to support the privacy protecting back-propagation algorithm on the ciphertexts. Zhang et al. \cite{Zhang_2016} adopted a fully homomorphic property BGV cryptographic system \cite{Zvika_2012} to carry out the computation through outsourcing. Among the homomorphic methods, fully-homomorphic encryption (FHE) has the ability to provide remarkable privacy. However, FHE introduces significant computational overhead and typically requires large key sizes as well as robust key management system to achieve sufficient security levels. Moreover, FHE algorithms are complex and challenging to implement correctly \cite{Shagufta_2020}. On the other hand, training on encrypted data leads to low prediction performance in practice  \cite{Li_Ping_2018}. Few researchers also used data obfuscation techniques to tackle data privacy \cite{pham2019data}. Obfuscation means adding or transforming data into a form that makes it more difficult to infer the original data samples. The obfuscation method adds random noise into the original samples or performs data masking with the original samples in such a way that the data becomes unusable for an attacker or an unauthorized personnel. Though obfuscation provides data privacy in a few cases, it loses the data characteristics which leads to a prediction trade-off. Another common way to mitigate the privacy preserving data sharing of sensitive user data is to de-identify the individual records. However, it is well-known that de-identified records can be easily re-identifiable by linking them to other identifiable datasets \cite{Jordon2019PATEGANGS} \cite{Emam_2011} \cite{Erlich_2014}. 

To build a machine learning model, real data is not always necessary. To mitigate the need of real data, synthetic data has been used for different purposes. Synthetic data can be generated from real data based on a concept called, differential privacy, using Generative Adversarial Networks (GANs) \cite{Goodfellow_2014} and differential private Wasserstein generative adversarial networks (DPWGANs) \cite{Mulder2019MSC}. And then we can say that this synthetic data is differentially private \cite{Dwork_2014} with respect to the original dataset. The differential privacy (DP) proposed by Dwork et al. \cite{Dwork_2006} in 2006 has shown provable privacy guarantees for individual records. Differential privacy (DP) is a strong, mathematical definition of privacy in the context of statistical and machine learning analysis \cite{Alexandra_2018}. DP is a criterion of privacy protection that analyzes sensitive personal information. The differentially private synthetic data ensures privacy to a large extent from the cloud service provider as well as assures that the adversaries are unable to assume any information about a single individual sample with high confidence from the output results of the machine learning classifiers. 

GANs is a powerful way to generate synthetic data but it does not provide any rigorous privacy guarantees. Therefore, in this study, synthetic data has been generated using DPWGANs (a version of GANs) which provides differentially private data. After generating the synthetic data, adaptive Weight-of-Evidence (aWOE) has been applied to it which adds another privacy layer to the datasets as well as improves the prediction performance. To the best of our knowledge, GANs based privacy preserving customer churn prediction has not yet been studied in the literature. This encourages us to test the effect of GANs on data privacy and prediction accuracy for customer churn prediction in the context of the telecommunication industry and this study shows several positive outcomes. In particular, this paper makes the following key contributions:
\begin{itemize}
       
    \item We have applied DPWGAN to generate synthetic data for the training of a CCP model in a privacy preserving manner.
    
    \item We have proposed an adaptive Weight-of-Evidence (aWOE) data transformation method that enhances prediction performance. We have applied the aWOE method on the synthetic data before feeding the data into the machine learning classifiers.

    \item We have trained eight different classifiers using synthetic data generated from our proposed GANs-aWOE framework. The classifiers we employed are Na\"{\i}ve Bayes (NB), Logistic Regression (LR), K-Nearest Neighbor (KNN), Random forest (RF), Decision tree (DT), Gradient boosting (GB), Feed-Forward Neural Networks (FNN), and Recurrent Neural Networks (RNN). The models have been compared against each other, as well as the state-of-the-art methods on three distinct publicly accessible datasets using a range of information retrieval metrics, including accuracy, specificity, precision, recall, F-measure, and AUC. The outcomes are very promising, and the privacy analysis indicates that the GANs-aWOE technique also offers data privacy assurances.
     
    \item  Thus we have put forth a privacy preserving customer churn prediction model (PPCCP) that preserves data privacy as well as improves prediction performances. To the best of our knowledge, this is the first study on data privacy based CCP model in the telecommunication industry.
 
\end{itemize}

The rest of the paper is organized as follows. Section \ref{sec:literature_review} presents the literature review. The materials and methods that were used in this study are presented in Section \ref{sec:Materials_Methodology}. Specifically, Subsection \ref{subsec:mWOE} describes our proposed adaptive Weight-of-Evidence (aWOE), while GANs-aWOE based privacy preserving CCP model is presented in subsection \ref{subsec:GANs-mWOE_CCP}. Section \ref{sec:Results} showcases the experimental results and performance comparisons. The discussion section is provided in Section \ref{sec:Discussion}. Finally, we conclude the proposed study with directions to future research in section \ref{sec:Conclusions}.

\section{Literature review} \label{sec:literature_review}
Within this section, we provide a concise overview of recent research works on privacy-preserving data mining in cloud environments. We also examine the strategies employed to mitigate the challenges posed by the CCP problem. 

Most of the privacy-preserving research works in machine learning are based on obfuscation, encryption, Homomorphic Encryption (HE), and differential privacy. Arockiam et al. in \cite{Arockiam2014} proposed a novel technique to achieve data security. The aim of the authors was to provide confidentiality using obfuscation and encryption. The authors in \cite{Hitendra2016} discussed obfuscation as a security measure in the cloud, and proposed a semi-encrypting technique that reduces the privacy risk. Data confidentiality issues in the cloud are presented by Manikandasaran et al. in \cite{Manikandasaran2016}. In \cite{Suthar2015}, the methodology for achieving data confidentiality, security, and integrity is discussed. In this scheme, encryption is primarily carried out at the client side, while obfuscation is provided by the cloud service providers. 

Li et al. in \cite{PingLi2018} presented a framework for preserving privacy during outsourced classification in the cloud, involving encrypted data and distinct public keys. To enable the computation over the encrypted data,  the encryption scheme should rely on a homomorphic property. To realize the computation on encrypted data, several homomorphic encryption schemes have been proposed in \cite{Fousse2011} and \cite{Paillier_1999}. Nonetheless, these are partial homomorphic encryption (PHE) schemes and can perform only a restricted set of operations (either addition or multiplication, but not both). Due to the lack of variety of arithmetic operations for practical applications, those PHE encryption schemes fail to meet the purpose. To address these limitations, researchers adopted fully homomorphic encryption (FHE) scheme,  which is capable of facilitating a diverse range of arithmetic computations concurrently. The FHE possesses the capability to directly execute computations on encrypted data as if it is executed on plaintext data. The first FHE scheme was presented by Gentry \cite{Gentry2009}. Next, various FHE schemes were proposed, like Brakerski et al. \cite{Brakerski2011,Brakerski2014}, Coron et al. \cite{Coron2011}, Jung et al. \cite{Jung2013}. However, those proposed FHE schemes are infeasible for use in extensive applications due to a common drawback: they involve computational burdens in relation to effectiveness and execution time that are not suitable for current cloud applications. As a result, it has been concluded that FHE lacks of necessary efficiency for real-world applications, particularly within the context of cloud computing.  

To tackle the data privacy in cloud and machine learning models recent researchers are using synthetic data (see Bindschaedler et al. \cite{Bindschaedler2017} and Huang et al. \cite{Huang2017} for example). In \cite{Bindschaedler2017}, Vincent et al. used a graphical probabilistic model to transform real data points into synthetic data points. Huang et al. \cite{Huang2017} introduced GANs with differential privacy (DP) to obfuscate the real data points and enable the privacy for more realistic datasets. To generate synthetic data, GANs are also used in \cite{Beaulieu2019,xie2018differentially,zhang2018differentially}. However, it has been proved that finding a stable set of training data with the necessary amount of noise is quite difficult. DPGAN is a modified GAN framework presented in \cite{xie2018differentially}, where the fundamental concept is to introduce noise to the discriminator's gradient during training, thereby establishing guarantees for differential privacy. These ideas are also used in \cite{Beaulieu2019}. Martin et al. \cite{Arjovsky2017} developed Wasserstein generative adversarial networks (WGANs) which are easier to train than traditional GANs. Based on the WGANs, Mulder \cite{Mulder2019MSC} proposed differential private Wasserstein generative adversarial networks (DPWGAN) to provide differential privacy guarantees of the generated synthetic data. In this study, we have used DPWGAN to generate the synthetic data which is used to train the CCP models.

To address the CCP problem, numerous Machine Learning (ML) and data mining strategies have been proposed including Rough set theory (RST) \cite{b12_Amin_Anwar, b13_Amin_Shehzad}, data certainty \cite{AMIN_2019_BR}, Naïve Bayes (NB) and Bayesian network (BN) \cite{b14_Kirui}, Decision tree (DT) \cite{b18, b19}, Logistic regression (LR) \cite{b19}, RotBoost (RB) \cite{Idris_Khan_b19}, Support Vector Machine (SVM) \cite{Renjith_b19}, Genetic algorithm based neural network \cite{b15_Pendharkar}, AdaBoost \cite{b16}, etc.  within the telecommunications industry (TCI), employing customer relationship management (CRM) data. The utilization of CRM data is prevalent in prediction and classification tasks \cite{b17}. However, the CRM data contains personal information and training models in the cloud, or sharing the CRM data with a third party creates privacy threats. To tackle this issue, in this study, we generate synthetic data using the previously mentioned DPWGAN, and the synthetic data has been used to train the model.

\section{Materials and Methodology} \label{sec:Materials_Methodology} 
\subsection{Datasets} \label{sec:Datasets} 
We utilize three publicly available benchmark datasets that have been widely examined to address the CCP challenges in the telecommunication area. The sample sizes of the dataset-1, dataset-2, and dataset-3 are $100000$, $7043$, and $5000$, respectively. More details about these datasets are presented in Table \ref{table:dataset}.

\begin{table}[h!]
\caption{Summary of datasets}
\label{table:dataset}
\begin{center}
\begin{tabular}{ p{5cm} p{2cm}  p{2cm}  p{2cm} }
 \hline
 \vspace{.05mm}\\
 Description& Dataset-1 &Dataset-2 & Dataset-3\\
 \vspace{.05mm}\\
 \hline
 \vspace{.01mm}\\
 No. of samples   & 100000    &7043 & 5000 \\
 No. of attributes&   101  & 21 & 20 \\
 No. of class labels & 2 & 2 & 2\\
 Percentage of churn samples    & 49.56 & 26.54 &  14.14  \\
 Percentage of non-churn samples &  50.43  & 73.46 & 85.86\\
 Source of the datasets& Kaggle \cite{Dataset_1_2022}     & Kaggle \cite{Dataset_2_2022}  & Data.world\cite{Dataset_3_2022}   \\
 \hline
\end{tabular}
\end{center}


\end{table}

\subsection{Preprocessing} \label{subsec:datapreparation} 
Data preprocessing stands as one of the challenging tasks for the data scientists. This technique guarantees that the format of the variables generated in the initial phase is well-suited for the churn prediction. Throughout the data preparation process, we executed the following key steps:

\begin{itemize}

 \item We ignore the attributes that represents unique identifiers and/or descriptive texts that serve only informational intentions.

 \item Redundant features have been eliminated.
 
 \item Based on previous studies \cite{COUSSEMENT201727} \cite{sana_JK_2022}, Numeric values that are missing are replaced with zero (0), while categorical values that are missing are treated as a distinct category.   
 
 \item Following our prior work \cite{sana_JK_2022}, categorical values of the form `yes' or `true' have been encoded as 1, while ‘no’ or ‘false’ have been encoded as 0. The numeric representation was achieved using the Label Encoder from the Python library sklearn. 
\end{itemize}

\subsection{Evaluation Measures} \label{subsec:evaluation_measure}
The confusion matrix is one of the commonly used performance measurements to evaluate the prediction performance of a model. For our proposed privacy prediction CCP model, the attributes of the confusion matrix have been outlined as follows: (i) True Positives (TP): accurately classified churn customers (ii) True Negatives (TN): accurately classified non-churn customers (iii) False Positives (FP): non-churn customers, erroneously classified as churned customers and (iv) False Negatives (FN): churn customers, erroneously classified as non-churn customers. The well-known evaluation metrics that are used to compare the performance of the models are presented in Table \ref{table:evaluation_measures}.

\begin{table}[h!]
\caption{List of evaluation measures}
\label{table:evaluation_measures}
\begin{tabular}{  p{2cm}  p{5cm}  p{8cm}  }
 \hline
 \vspace{.05mm}\\
 Metric & Description & Equation\\
 \vspace{.05mm}\\
 \hline
 \vspace{.01mm}\\
 Accuracy & It is the rate of accurate prediction of the model & 
 Mathematically Accuracy can be expressed as:
 \begin{equation} \label{eq:Accuracy}
       Accuracy = \frac{TP+TN}{TP+FN+FP+TN}
  \end{equation}  
  \\

Specificity & It is the accurate prediction rate of non-churn customers & 
 The mathematical equation of specificity is:
 \begin{equation} \label{eq:Specificity}
       Specificity = \frac{TN}{FP+TN}
  \end{equation}  
  \\
  
  Precision & It is the positive predictive value of the model & 
 Precision can be define as:
   \begin{equation} \label{eq:precision}
       Precision = \frac{TP}{TP+FP}
  \end{equation}
  \\

   Recall & Utilizing recall as an evaluation metric is a valid choice when the objective is to capture the maximum number of true churn customers. & 
 The equation of the recall is:
  \begin{equation} \label{eq:recall}
      Recall = \frac{TP}{FN+TP}
  \end{equation}
  \\

   F-Measure & The F-measure represents the harmonic mean of the precision and recall. F-measure is used to achieve a balance between precision and recall. & 
The Mathematical formula of F-measure is defined below. 

  \begin{equation} \label{eq:f-measure}
       \text{F-Measure} =\frac{(2*precision*recall)}{(precision+recall)}
  \end{equation}
  \\

   AUC &  AUC refers to the area under the receiver operating characteristic (ROC) curve which represents the trade-off between the true positive rate and false positive rate. &

   AUC is not represented by a specific equation. Instead, it is calculated by numerically integrating the ROC curve. 
  \\
  
 \hline
\end{tabular}
\end{table}

\subsection{Generative Adversarial Networks (GANs)} \label{subsec:GANs} 
Generative adversarial networks (GANs) are generative models proposed by Goodfellow et al. \cite{Goodfellow_2014}. Since the introduction of GANs, they are attracting growing interest in many applications. GANs are generative models that generate data by inherent features. GANs were designed to be able to learn a complex distribution and then sample from them. DCGAN \cite{radford2015unsupervised} and WGAN \cite{MartinArjovsky2017} are the two significant milestones in the evolution of GANs. In contrast to earlier GANs, DCGAN model offers greater controllability and is less susceptible to collapse. Diverse GAN models have been successively developed based on different needs and found widespread use in various applications such as computer vision, medical, art, and security encryption fields. The main advantage of WGAN is that it gives more stable training process and less sensitive to hyperparameter and model architecture. BigGAN \cite{brock2018large} shows better performance in generating images characterized by superior fidelity and a narrow variety gap. StyleGAN \cite{Karras2019} has shown a remarkable progress in facial generation tasks. AgecGAN \cite{Sheng2020} proved its capability in the field of cross-age face recognition, the search for missing children, and entertainment applications. GANs based systems are also developing rapidly in many other fields like image processing, video processing, speech processing, text processing, signal processing, ECG, and EEG signal recognition, etc. \cite{Lianchao2020}. Therefore, GANs and their various diverse derivative models possess broad significance. The research and application of GANs are currently in a rising phase, indicating a promising scope for future development and widespread application.

In a GAN, there are two fundamental elements: the generator and the discriminator. These components are two neural networks, one designed to create data and the other trained to distinguish fake data samples from genuine data (hence the ``adversarial'' nature of the model) \cite{LeCun2015}. Generators are able to generate new data samples from real data distribution and a little bit of random noise. These generated synthetic samples are fed into the discriminator. The discriminator is simply a classifier and can discriminate between real data and fake data. The discriminator penalizes the generator for generating doubtful results and updates the discriminator's parameters to minimize its classification error. At the outset of training, the generator generates fake data, and the discriminator swiftly learns to recognize that it is fake. Finally, if the generator's training goes fine, the discriminator gets worse at identifying the real and fake samples. It starts to identify the fake data as real, and its accuracy declines. 

\subsection{Differential Privacy (DP)}  \label{subsec:DP} 
The differential privacy (DP) \cite{DworkCynthia2006} is a protective way of identity and privacy of individual records in a population distribution. DP is not an algorithm; instead, it encompasses a set of criteria that must be fulfilled to ensure that sharing supplementary or partial information will not lead to privacy violation. As a common practice, differentially private techniques uphold the confidentiality of individual data by incorporating random noise (usually adding or removing a few records) when producing statistics that does not influence the outcome of any useful analysis. DP can create an obstacle for an adversary with supplementary information and can provide effective privacy protection. The fundamental concept of DP is that if an adversary can not distinguish between a dataset that includes a specific record and a dataset that lacks of that record, the attacker will be incapable of deducing any further information. Differential privacy provides a balance between privacy and utility by modulating a privacy budget parameter, where a smaller privacy budget index provides more robust privacy guarantees. In order to extract valuable information from a dataset without jeopardizing privacy about a particular value, a condition needs to be satisfied. The condition is that any query on a dataset where at most one data sample has been changed should not result in the ability to infer any additional information for a particular data point. It's important to note that the outcome of any query executed on the modified dataset must not be distinguishable from the outcome of the same query conducted on the real dataset. The definition of differential privacy was first proposed by Cynthia Dwork \cite{DworkCynthia2006}, who showed that DP can provide strong privacy guarantee. In \cite{DworkCynthia2006}  the definition of ``Differential Privacy'' ($\epsilon$-DP ) is defined as follows. 

Definition : Let $\epsilon \geq 0$. A mechanism M is considered to be $\epsilon$-DP if for every pair of neighboring databases $D_1$ and $D_2$ that differ in at most one element, and all $E \subseteq Range(M)$,

 \begin{equation}  \label{eq:dp}
      P[ M(D_1) \in E] \leq \exp (\epsilon) \times P [ M(D_2) \in E ]
 \end{equation} 
 
Where E is the all possible outputs set of M. M(D) is the privacy preserving mechanism. $P$ is the probability distribution of M(D). The value of $\epsilon$, represents the privacy index that controls the privacy level: lower $\epsilon$ value achieves better privacy protection \cite{Lee2011, xie2018differentially}. In this mechanism, the random noise must be sampled from a probability distribution that aligns with the stipulated criteria. This noise is subsequently introduced into the query result.  The outcome generated through this mechanism is proportional to the noise required to ensure differential privacy. The magnitude of the noise must be significant enough to be able to cover any changes made. As a result, any change applied to the dataset will lead to a change in the query's output.  Laplace mechanism \cite{Geng_2016}, Gaussian mechanism \cite{Jinshuo2022}, Exponential mechanism \cite{Frank2007} and Private Aggregation of Teacher Ensembles (PATE) \cite{Jordon2019PATEGANGS} are the most commonly used  mechanisms for differential privacy. Among these, Gaussian mechanism provides smoother noise distribution and more flexibility in noise calibration and is thus widely used in privacy-preserving algorithms \cite{Canonne_Kamath_Steinke_2022, xie2018differentially, Dwork2014}. Therefore, in this research we used Gaussian mechanism based differential privacy.

\subsection{Training GANs With Differential Privacy}  \label{subsec:GANs_DP} 
Differential privacy ensures that the distribution of an algorithm's output remains almost unaffected by the inclusion or exclusion of an individual record. By analyzing the outputs, an adversary is unable to get additional information about individuals, which ensures privacy. In the context of GANs, differential privacy holds the potential to facilitate precise learning of the data distribution, even when adding or removing training samples.

To avoid leaking sensitive information of individuals, recently researchers are using synthetic data which are generated using GANs. Usually, GANs are trained with differential privacy guarantees. The approach introduced by Abadi et al. \cite{Abadi2016}, known as differentially private stochastic gradient descent (DPSGD), uses noisy stochastic gradient descent to mitigate the impact of individual training samples. PATE-GAN \cite{Jordon2019PATEGANGS} was introduced to create synthetic multi-variate tabular data while preserving the confidentiality of the training data. DP-CGAN was developed by Torkzadehmahani et al. \cite{DPCGAN2019} to produce synthetic data along with their corresponding labels. The main objective of SPRINT-GAN \cite{Beaulieu2019} is to distribute patient-level clinical trial data with differential privacy measures. This guarantees that the synthetic data does not disclose the identities of trial participants. Augenstein et al. \cite{augenstein2019generative} developed differentially private GANs models through federated learning. This means that the raw data is spread out across various user devices, and a central server is responsible for coordinating the training of a shared global model. A very recent method, DP-WGAN developed by Mulder \cite{Mulder2019MSC} is suitable for tabular data and achieves differential privacy in GANs by cautiously adding noise to gradients during the learning process. The amount of noise added depends on privacy privacy budget parameter $\epsilon$. A smaller $\epsilon$ will yield better privacy but less accurate response \cite{Lee2011, xie2018differentially}. In this study, we use the DP-WGAN to generate the synthetic data where we set privacy budget $\epsilon=10$ as suggested and experimentally shown in different studies like \cite{Brett_2018, xie2018differentially, Yang_2020}.

\subsection{Adaptive Weight-of-evidence (aWOE)} \label{subsec:mWOE} 
The Weight-of-Evidence (WOE) \cite{COUSSEMENT201727} technique involves binning and log transformation. In most cases, WOE can solve the issue of skewed data distribution. Kristof et al. \cite{COUSSEMENT201727} and our previous study \cite{sana_JK_2022} show that WOE captures important patterns of the dataset and improves the prediction performance. It is basically the natural logarithm (ln) of the ratio between the distribution of good events (1) and the distribution of bad events (0). The equation of the Weight-of-evidence (WOE) for each group is given bellow.

\begin{equation} \label{eq:Weightofevidence}
       WOE = ln\bigg(\frac{\text{Percentage of churn customers in a particular group}}{\text{Percentage of non-churn customers in a particular group}}\bigg)
\end{equation}

\begin{figure}[!htb]
\begin{center}
\includegraphics[height=450px,width=300px]{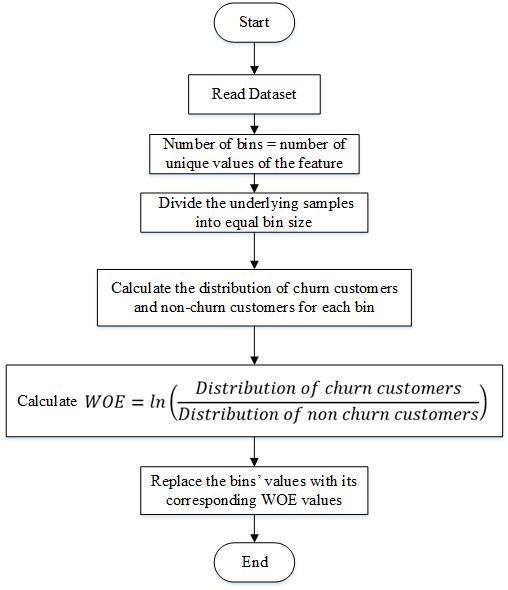}
\caption{Flowchart of Weight-of-Evidence (WOE), implemented in the mlencoders library.}
\label{fig:WOE_flow_chart}
\end{center}
\end{figure}

\begin{figure}[!htb]
\begin{center}
\includegraphics[height=450px,width=325px]{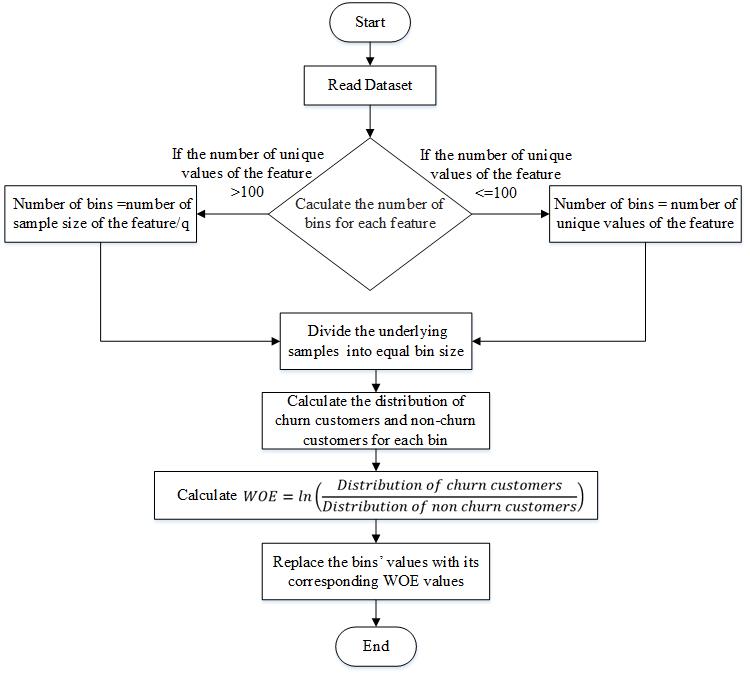}
\caption{Flowchart of Adaptive Weight-of-Evidence (aWOE).}
\label{fig:mWOE_flow_chart}
\end{center}
\end{figure}

Although WOE is very effective and improves the prediction performance, an appropriate number of bins ($b$) can achieve more improved prediction performance. In this study, we propose adaptive weight-of-evidence (aWOE) which uses dynamicnumber of bins and achieves superior prediction performance. In the convention WOE method (implemented in the Python programming language library mlencoders), the number of bins ($b$) is equal to the number of unique values of the data feature. Our proposed aWOE behaves identical to the vanilla WOE when the number of unique values is less than or equal to 100. When it is greater than 100, on the other hand, we determine the number of bins by dividing the number of samples by $q$, where $q$ is a tunable parameter. Additionally, when a particular bin does not contain any event or non-event, WOE ignores the value which affects the prediction performance. Therefore, in Equation~\ref{eq:Weightofevidence}, we have added an adjustment constant of 0.0001 in both numerator and denominator. The Figure \ref{fig:WOE_flow_chart} and \ref{fig:mWOE_flow_chart} show the flowcharts of the conventional WOE (implemented in mlencoders library) and our proposed adaptive counterpart, respectively.

\begin{table}[h!]
\caption{List of baseline classifiers.}
\label{table:BaseLineClassifiers}
\begin{tabular}{ p{1cm} p{2.5cm}  p{3cm}  p{8cm}  }
 \hline
 \vspace{.05mm}\\
 Key& Classifer & Model type& Description\\
 \vspace{.05mm}\\
 \hline
 \vspace{.01mm}\\
 NB   & Naïve Bayes & Gaussian  &  NB is a conditional probabilistic learning algorithm. It is based on the Bayes theorem \\
 KNN   & K-Nearest Neighbor & Distance based learning &  KNN algorithm calculates similar things based on the nearest neighbors.\\
 LR & Logistic Regression & Statistical model&  LR is the form of binary regression. It estimates the probability of an event occurring.\\
 RF&    Random forest  & Trees   & RF is a tree-based ensemble learning algorithm\\
 DTree&    Decision tree   & Trees   & DT is a tree structure based classification or regression model.\\
 GB&   Gradient boosting   & Trees   & GB is an ensemble tree-based (usually decision trees) boosting technique.\\
 FNN&   Feed-Forward Neural Network & Deep learning   & The FNN is a classifier that utilizes deep learning and processes input data in a one-way direction.\\
 RNN&    Recurrent Neural Network & Deep learning   &  The RNN is a type of artificial neural network designed to work with sequence data, such as time series or natural language. It has connection that loop back on it, allowing it to maintain a memory of previous input. \\
 \hline
\end{tabular}
\end{table}

\subsection{Privacy Preserving GANs-aWOE CCP Framework} \label{subsec:GANs-mWOE_CCP} 
\begin{figure}[!htb]
\begin{center}
\includegraphics[height=400px,width=250px]{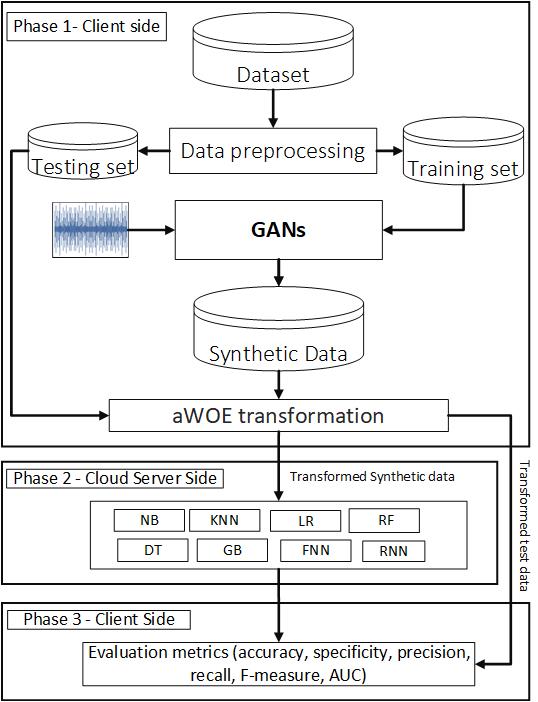}
\caption{Comprehensive block diagram of GANs-aWOE based Privacy-Preserving CCP Framework}
\label{fig:GANs_mWOE_framework}
\end{center}
\end{figure}

Figure \ref{fig:GANs_mWOE_framework} outlines the comprehensive block diagram of the proposed PPCCP framework, which combines the GANs with aWOE for privacy preservation.  First, we performed essential preprocessing tasks on the datasets. Then, synthetic data has been generated using GANs. To train the GANs, we used 70\% data of the dataset and we kept separate 30\% of the original data for the model testing. To generate the privacy protected synthetic data, we trained the GANs (DP-WGAN) with the privacy buget $\epsilon =10$ as per the differential privacy condition and recommendation of \cite{Yang_2020}, \cite{Brett_2018} \cite{xie2018differentially}. Next, aWOE transformation method has been applied on the synthetic data. After that, the transformed synthetic datasets are used to train the machine learning classifiers. Finally, in the phase-3, the classifiers have been tested using the 30\% original test datasets (transformed using aWOE) and we measured the prediction performance in terms of accuracy, specificity, precision, recall, F-measure, and AUC (discussed in section \ref{sec:Results}). To test the performance of our proposed model, after the GANs training process, we generate 10 different synthetic datasets for each original dataset and we repeat the testing phase 10 times and report the average performance results. The steps of phase 1 are implemented in the client premises, the steps of phase 2 are done in the cloud server and the steps of the phase 3 are executed at the client side.

\subsection{Coding and Experimental Environment}
The experiments were carried out on a machine having Windows Server 2019, 64-bit system with Intel Xeon Silver 4214R 2.4GHz CPU, 64GB RAM, and 1TB HD. The code was written in Python 3.7 and Jupyter Notebook was utilized for coding. We also used Google Colab (https://colab.research.google.com/) for training the GANs model. The complete set of data and code can be accessed through the following link:  {https://github.com/joysana1/PPCCP}.

\section{Results} \label{sec:Results} 
We rigorously test the performance of our proposed aWOE DT method and compare the performance of the aWOE with that of conventional WOE DT method. We also test the performance of aWOE method with different number of bins to find appropriate number of bins.

In order to find the performance of our proposed PPCCP technique, we experimented with eight classifiers namely: Naïve Bayes (NB), K-Nearest Neighbor (KNN), Logistic Regression (LR),  Random Forest (RF), Decision Tree (DTree), Gradient Boosting (GB), Feed-Forward Neural Network (FNN), and Recurrent Neural Networks (RNN). At first, RAW data based CCP models (baseline classifiers) were tested. Here, we did not use our proposed GANs generated synthetic data and aWOE transformation. In the next step, we performed the WOE on the GANs generated synthetic data and used these data for the model training. We called these models are privacy preserving CCP models. 

\subsection{Performance Comparison Between aWOE and WOE DT method} \label{sec:aWOE_WOE} 
\begin{figure}[!htb]
\begin{center}
\includegraphics[height=200px,width=250px]{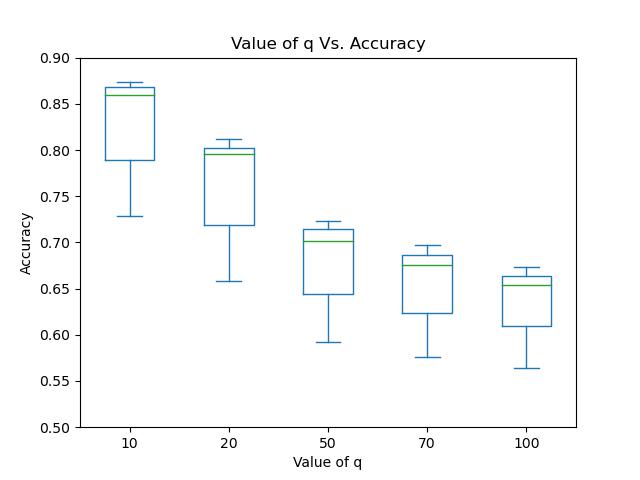}
\caption{Performance distribution of all classifiers for different $q$ values in terms of Accuracy on dataset-1}
\label{fig:aWOE-box-plot-accuracy}
\end{center}
\end{figure}

\begin{figure}[!htb]
\begin{center}
\includegraphics[height=200px,width=250px]{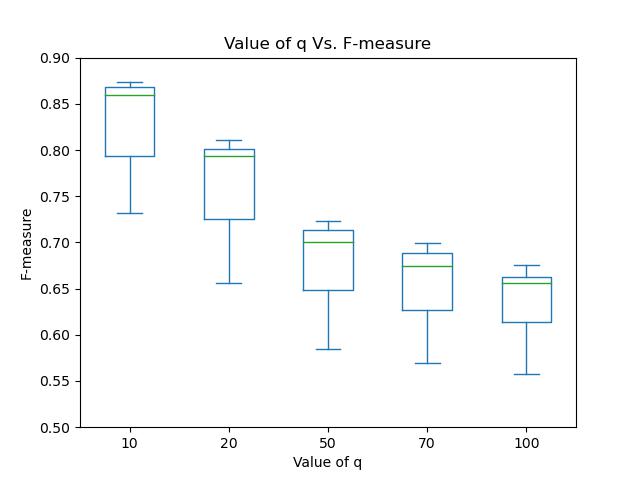}
\caption{Performance distribution of all classifiers for different $q$ values in terms of F-measure on dataset-1}
\label{fig:aWOE-box-plot-f1-score}
\end{center}
\end{figure}

The main purpose of the experiment is to determine whether our proposed aWOE method achieves better prediction performance than that of conventional WOE method. At first, we identify the optimal number of bins of our proposed aWOE then we compare the prediction performance between the aWOE and conventional-WOE methods. To find the optimal number of bins of our proposed aWOE method, in this experiment, we set the number of bins ($b$) is equal to the sample size divide by $q \in \{10, 20, 50, 70, 100\}$, when the number of unique values (category) of the data feature is greater than 100. Otherwise, the number of bins is equal to the number of unique values (category) of the data feature. Figure \ref{fig:aWOE-box-plot-accuracy} and \ref{fig:aWOE-box-plot-f1-score} illustrate the performance distribution of all the classifiers (used in this research) in terms of accuracy and F-measure across various $q$ values on dataset-1, respectively. The both figures clearly show the highest predictive performance for $q=10$. Therefore, this value was used for the rest of our experiments. 

\begin{figure}[!htb]
\begin{center}
\includegraphics[height=200px,width=250px]{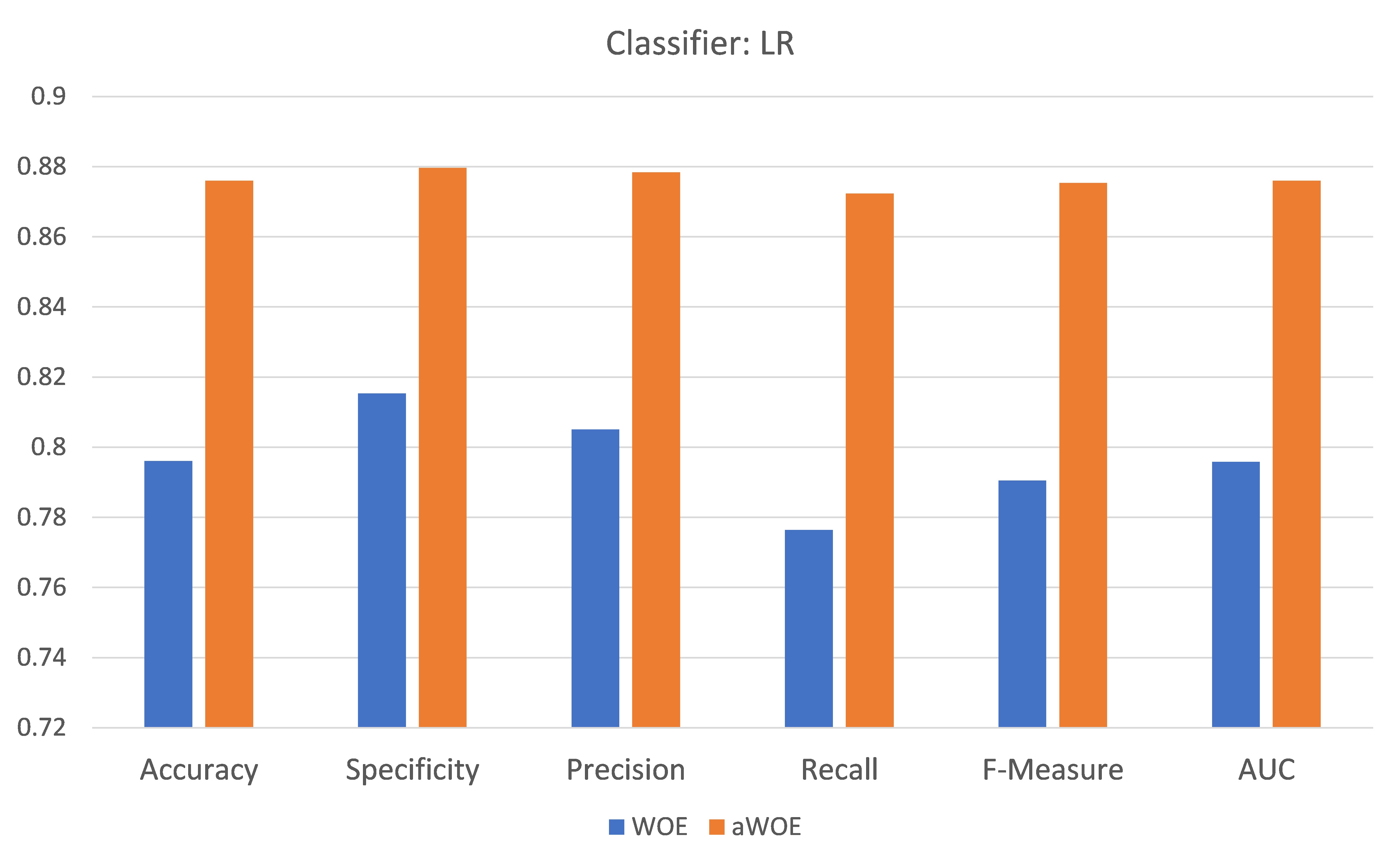}
\caption{Performance Comparison between aWOE and WOE method for LR classifier on dataset-1}
\label{fig:aWOE-WOE-LR}
\end{center}
\end{figure}

\begin{figure}[!htb]
\begin{center}
\includegraphics[height=200px,width=250px]{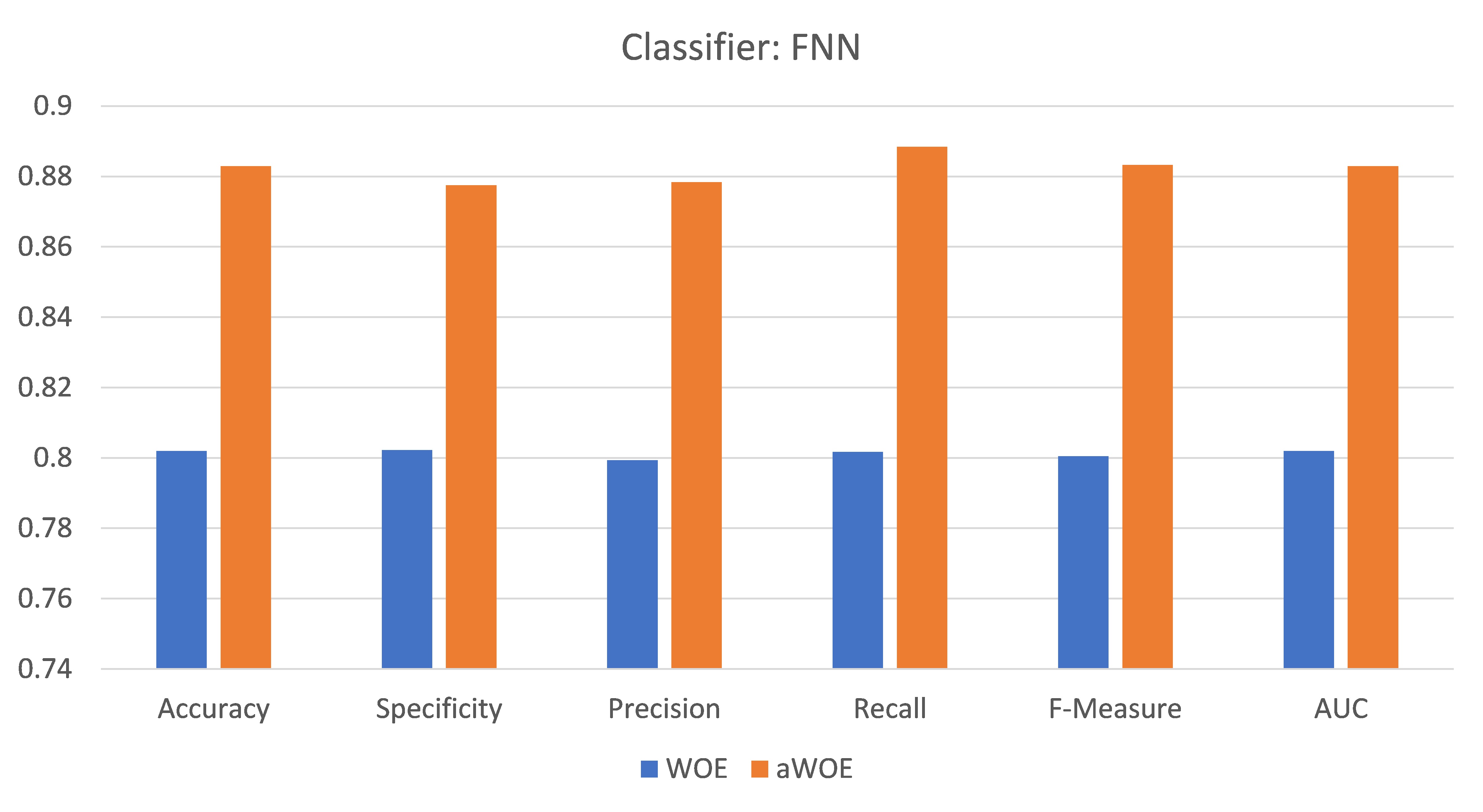}
\caption{Performance Comparison between aWOE and WOE method for FNN classifier on dataset-1}
\label{fig:aWOE-WOE-FNN}
\end{center}
\end{figure}

Figure \ref{fig:aWOE-WOE-LR} and \ref{fig:aWOE-WOE-FNN} represent the performance comparison between aWOE and WOE method on dataset-1 for classifier LR and FNN, respectively. We provided only the two classifiers results because they achieve relatively better result for both aWOE and WOE based models. For both classifiers, aWOE method outperforms the conventional WOE based transformation across the all performance metrics. Therefore, we use aWOE method for our proposed privacy-preserving customer churn prediction model.

\subsection{Experimental Results for Privacy-Preserving CCP model}
\begin{figure}[!htb]
\begin{center}
\includegraphics[height=200px,width=250px]{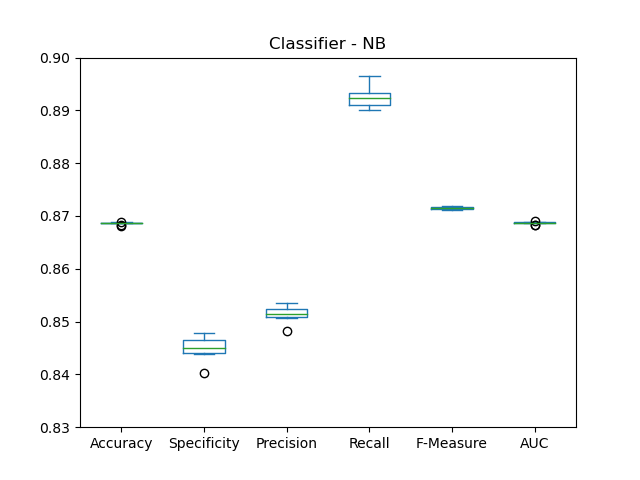}
\caption{Performance of NB classifiers trained on the 10 different synthetic datasets corresponding to original dataset-1.}
\label{fig:NB-10-synthetic datasets_performance}
\end{center}
\end{figure}

\begin{figure}[!htbp]
\begin{center}
\includegraphics[height=180px,width=220px]{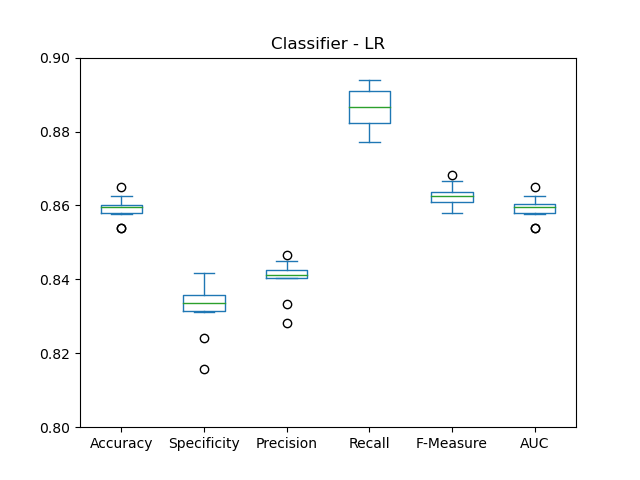}
\caption{Performance of LR classifier trained on the 10 different synthetic datasets corresponding to original dataset-1.}
\label{fig:LR-10-synthetic datasets_performance}
\end{center}
\end{figure}

To confirm the robustness of the GANs model for synthetic data generation, we generated 10 different training datasets from 70\% portion of each of the original datasets. On each such synthetic data, various models were trained, which were then assessed using the remaining 30\% original data. Figures \ref{fig:NB-10-synthetic datasets_performance} and \ref{fig:LR-10-synthetic datasets_performance} represent the box-plot of accuracy, specificity, precision, recall, F-measure, and AUC of NB and LR classifiers trained on the 10 different synthetic datasets corresponding to the original dataset-1. Those box-plots show that the performance is not sensitive to different generations of the synthetic data. For other classifiers, we also found the similar pattern. Figures \ref{fig:comparison_dataset_1}, \ref{fig:comparison_dataset_2} and \ref{fig:comparison_dataset_3}  represents the comparison among the RAW , GANs, aWOE and GANs-aWOE based CCP models for the dataset 1, 2 and 3 respectively. As mentioned earlier, for each original training dataset, 10 different synthetic datasets have been generated and and the results shown in these figures are averaged across the models trained with these datasets.

\subsubsection{Results on Dataset-1}
Across all metrics, the baseline classifiers (RAW) display poor performance on dataset-1 (Figure \ref{fig:comparison_dataset_1}). The highest F-measure is achieved by NB, but the value remains merely 0.633. The classifiers trained based on GANs synthesized data show inconsistent performance. In few cases, it shows high peak like recall in NB and specificity in RF, GB and RNN. In other cases, GANs based classifiers display very poor performance, like specificity in NB, recall in RF, GB, RNN etc. The best performance achieved by the GANs based classifier is 0.605 (for DTree) in terms of F-measure. These results indicate that the GANs based learning algorithms have failed to learn much. On the other hand, aWOE based classifiers and GANs-aWOE based classifiers consistently show better performance than RAW and GANs based classifiers. In few cases, aWOE based classifiers achieve highest performance, while in some other cases, GANs-aWOE  based classifiers achieve the best results. Only in two cases the RAW based classifiers (tree based classifiers like RF, DTree) are able to perform slightly better than GANs-aWOE based classifiers in terms of recall. Perhaps, the GANs-aWOE based transformation of the dataset does not affect much the data properties that are used for the tree-based classifiers. Hence, the performance of the tree-based classifiers are quite similar in both cases. Across all combinations in dataset-1,  the most remarkable achievement is attributed to aWOE based FNN with an impressive F-measure of 0.883. On the other hand, GANs-aWOE based NB classifier also achieves commendable performance (F-measure: 0.871).  In terms of accuracy and AUC, the GANs-aWOE based CCP classifiers outperform the RAW based classifiers for all the cases. 0.869 is the best accuracy value for GANs-aWOE based classifiers which is achieved by NB. 

\subsubsection{Results on Dataset-2}
\begin{figure}[h!]
    \centering
    \begin{subfigure}{0.3\textwidth}
        \includegraphics[width=\linewidth]{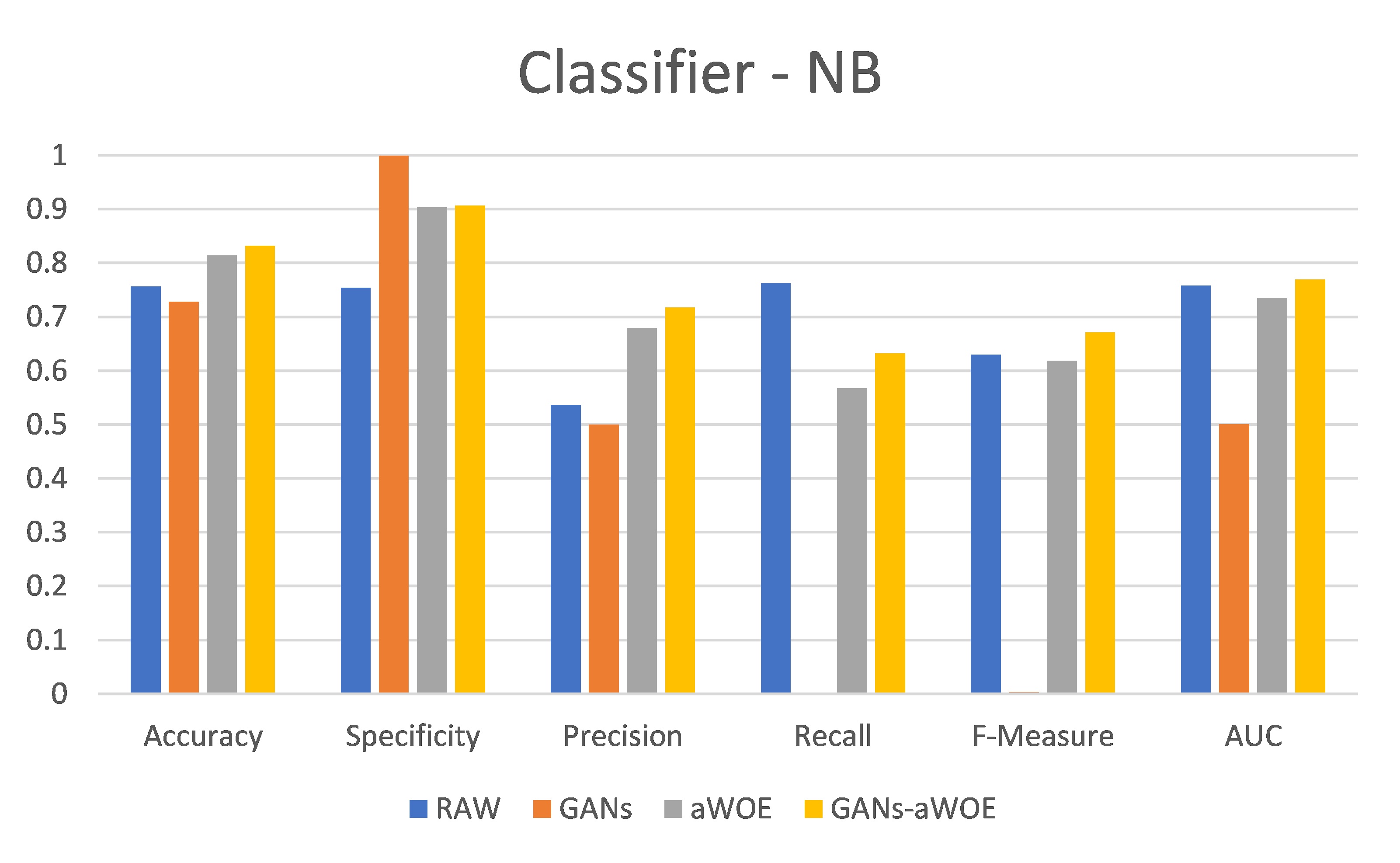}
        \caption{}
    \end{subfigure}
    \begin{subfigure}{0.3\textwidth}
        \includegraphics[width=\linewidth]{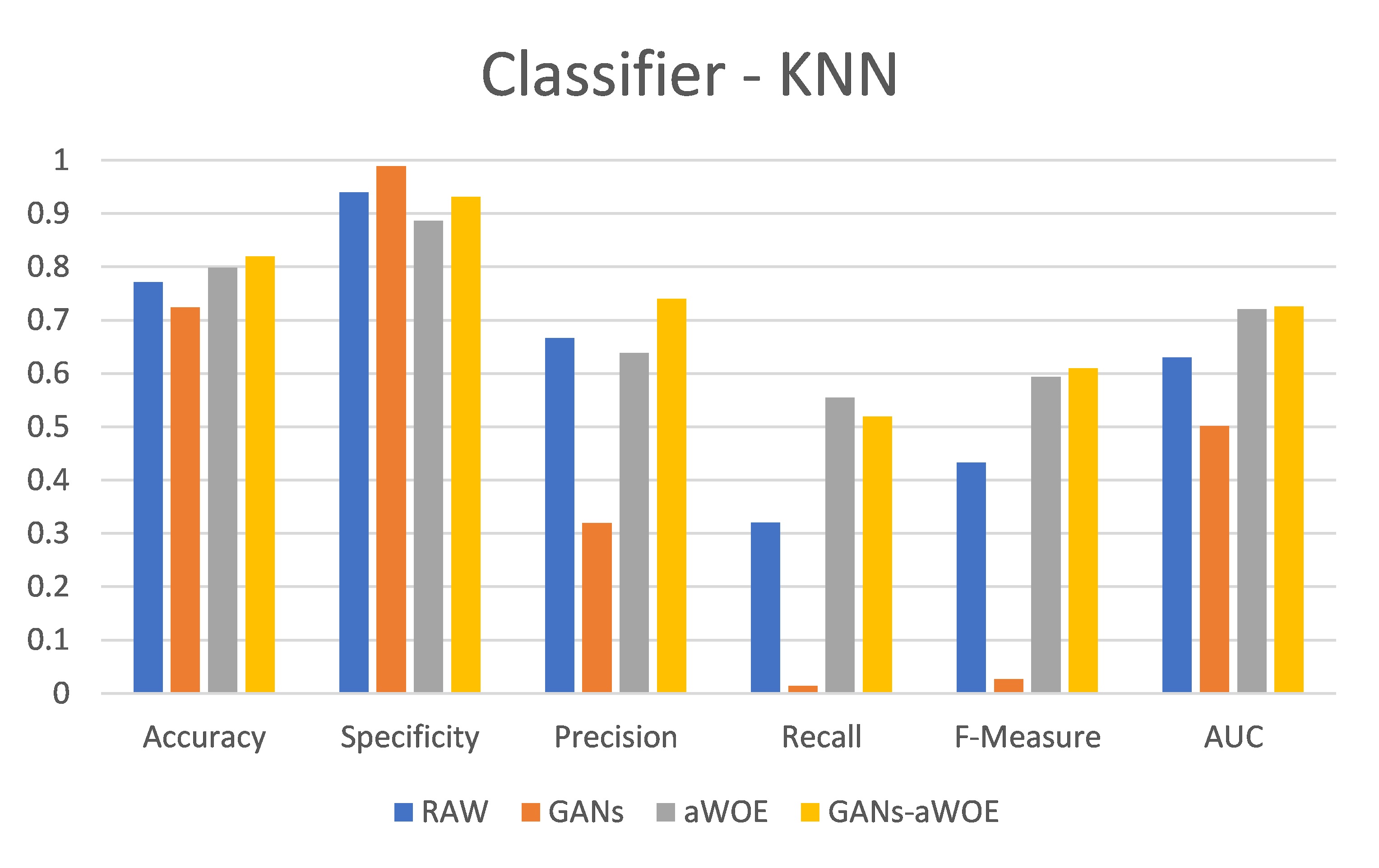}
        \caption{}
    \end{subfigure}
    \begin{subfigure}{0.3\textwidth}
        \includegraphics[width=\linewidth]{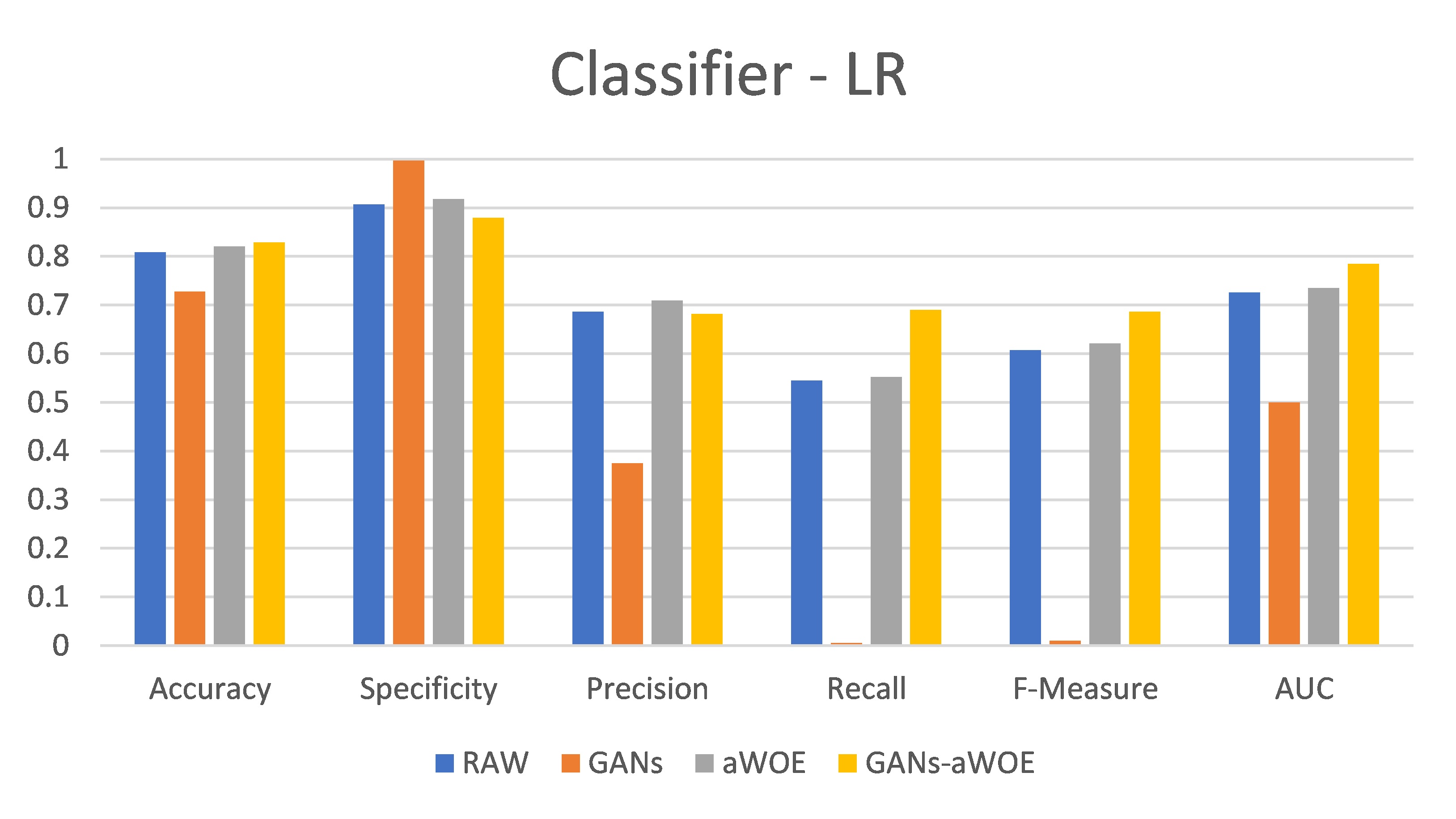}
        \caption{}
    \end{subfigure}
    
    \begin{subfigure}{0.3\textwidth}
        \includegraphics[width=\linewidth]{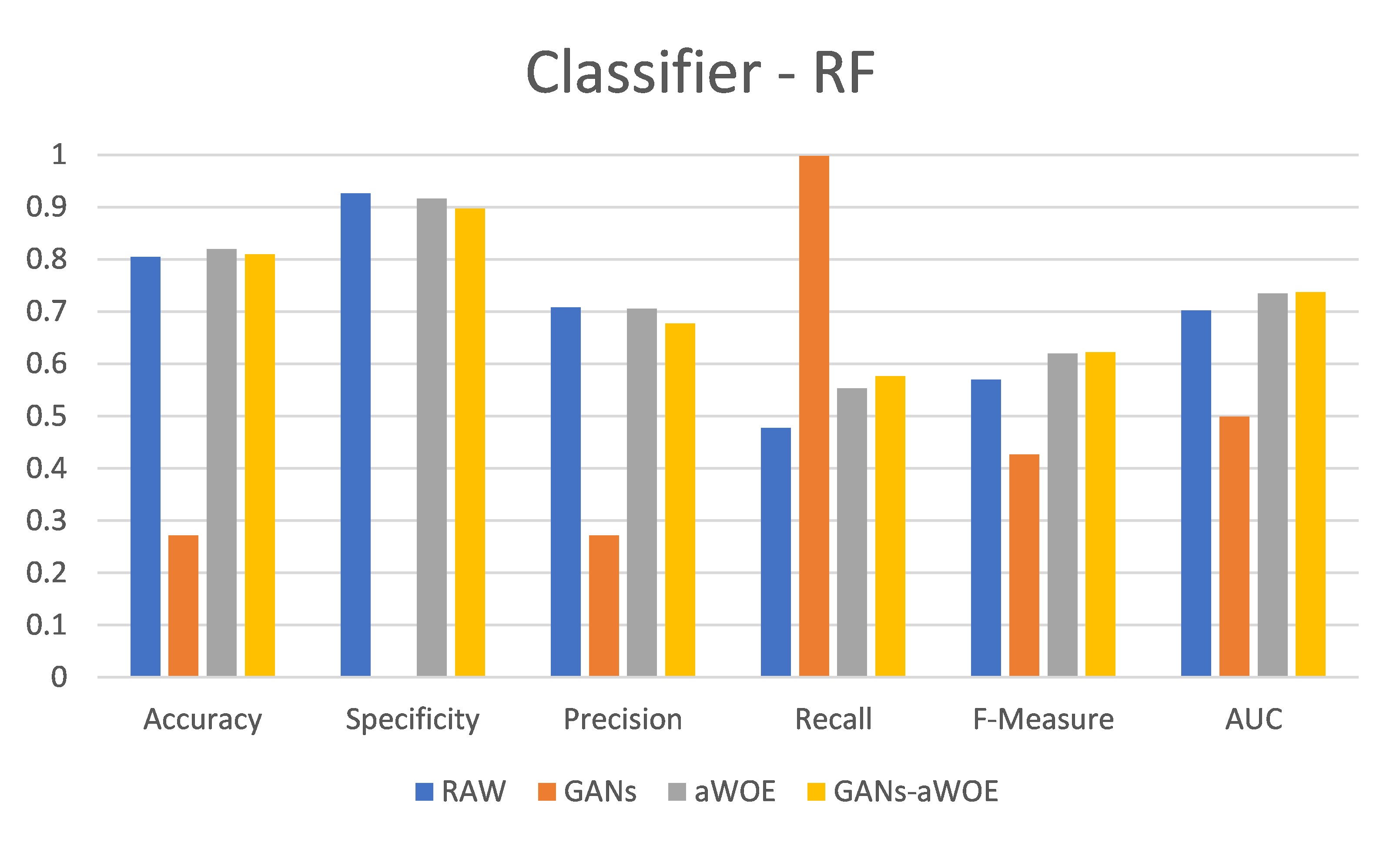}
        \caption{}
    \end{subfigure}
    \begin{subfigure}{0.3\textwidth}
        \includegraphics[width=\linewidth]{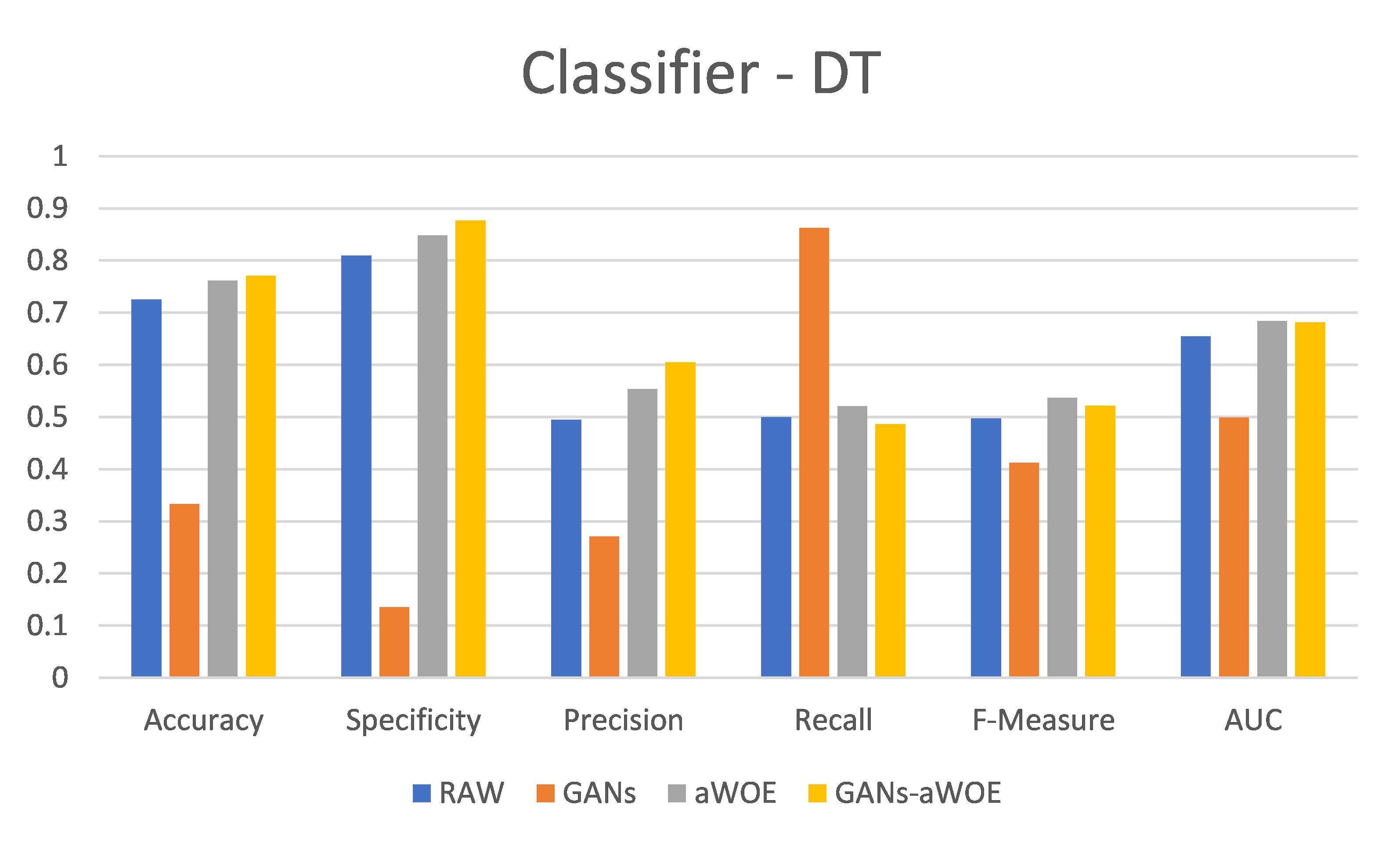}
        \caption{}
    \end{subfigure}
    \begin{subfigure}{0.3\textwidth}
        \includegraphics[width=\linewidth]{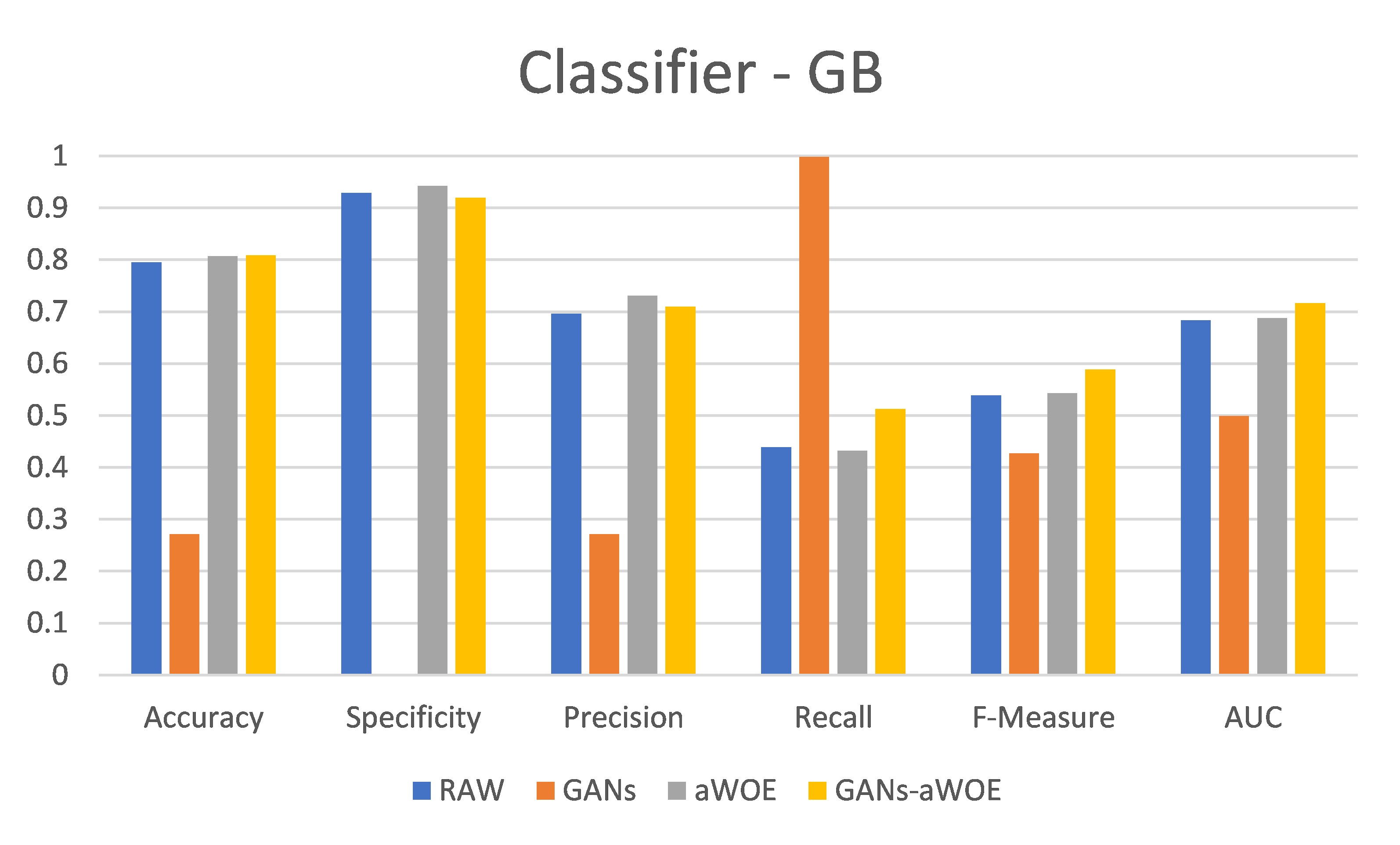}
        \caption{}
    \end{subfigure}

    \begin{subfigure}{0.3\textwidth}
        \includegraphics[width=\linewidth]{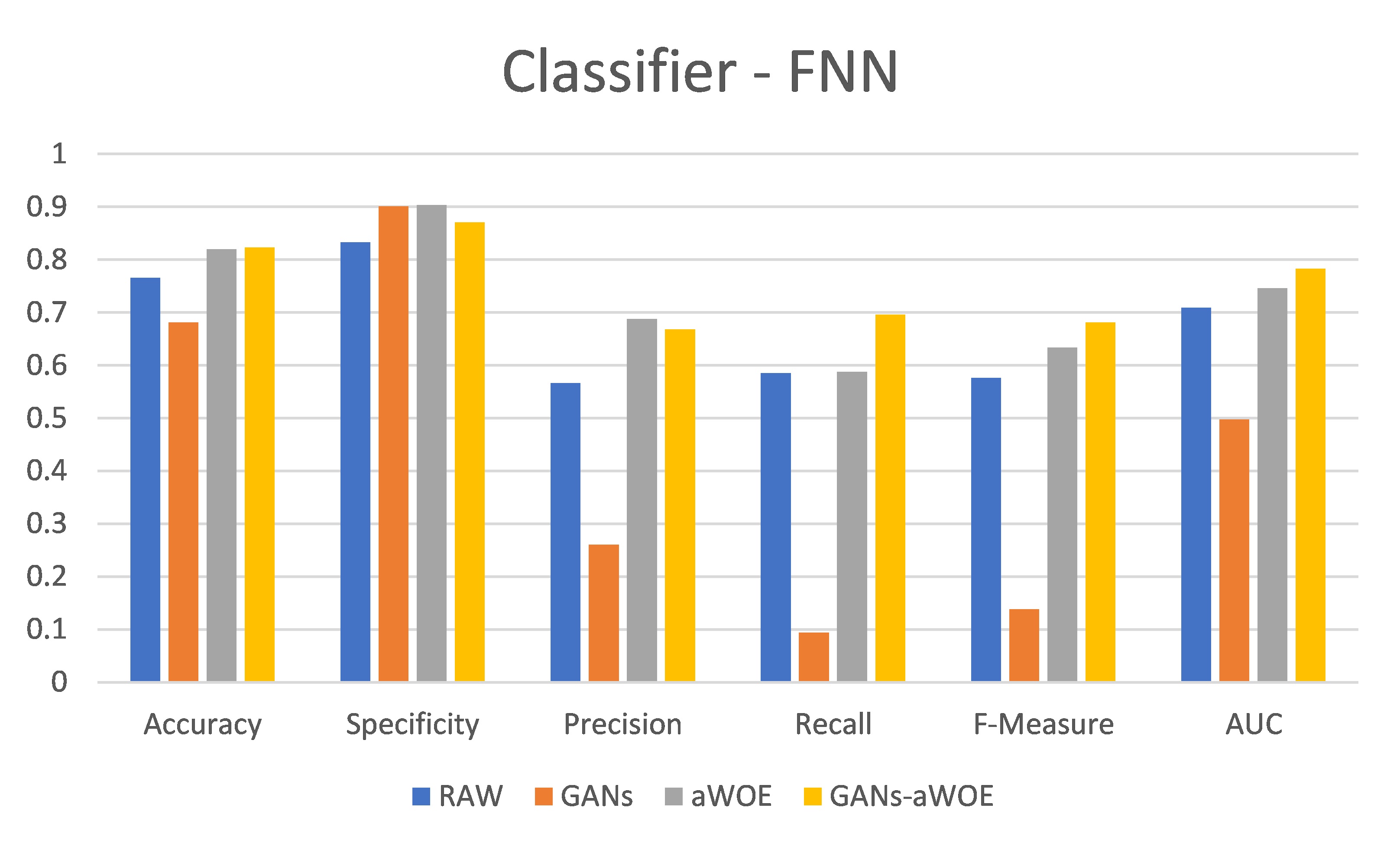}
        \caption{}
    \end{subfigure}
    \begin{subfigure}{0.3\textwidth}
        \includegraphics[width=\linewidth]{images/RNN-dataset-2.jpg}
        \caption{}
    \end{subfigure}
    \caption{Performance comparison among the RAW, GANs, aWOE and  GANs-aWOE based CCP models on dataset-2}
      \label{fig:comparison_dataset_2}
\end{figure}

We observed almost the same general performance pattern in dataset-2 as well. (Figure \ref{fig:comparison_dataset_2}). GANs based classifiers show inconsistent performance. The aWOE based classifiers and GANs-aWOE based classifiers show consistent and better performance than the baseline (RAW) classifiers and GANs based classifiers. Except the RNN, for all other cases the GANs-aWOE based classifiers outperform the RAW based classifiers in terms of accuracy. In terms of F-measure, the GANs-aWOE based classifiers achieve higher performance than the RAW based classifiers for all cases except the RNN. In terms of recall, the GANs-aWOE based classifiers also show higher performance except for NB and DTree classifiers. The other few exceptions we found were for KNN, LR, RF, and GB classifiers in terms of specificity, where GANs-aWOE performs slightly worse than RAW. In terms of AUC, GAN-aWOE based classifiers outperform the other techniques except the RNN. In all cases, aWOE based classifiers outperform the RAW and GANs based classifiers in term of accuracy. In terms of AUC, expect the RNN, for all other cases, the aWOE based classifiers achieve better performance than the RAW and GANs based classifiers.  On dataset-2, the highest performance score attained by the GANs-aWOE based NB classifier is 0.832 for accuracy, while for RAW, it stands at 0.756. The GANs-aWOE based LR achieves the best AUC value which is 0.785.

\subsubsection{Results on Dataset-3}
\begin{figure}[h!]
    \centering
    \begin{subfigure}{0.3\textwidth}
        \includegraphics[width=\linewidth]{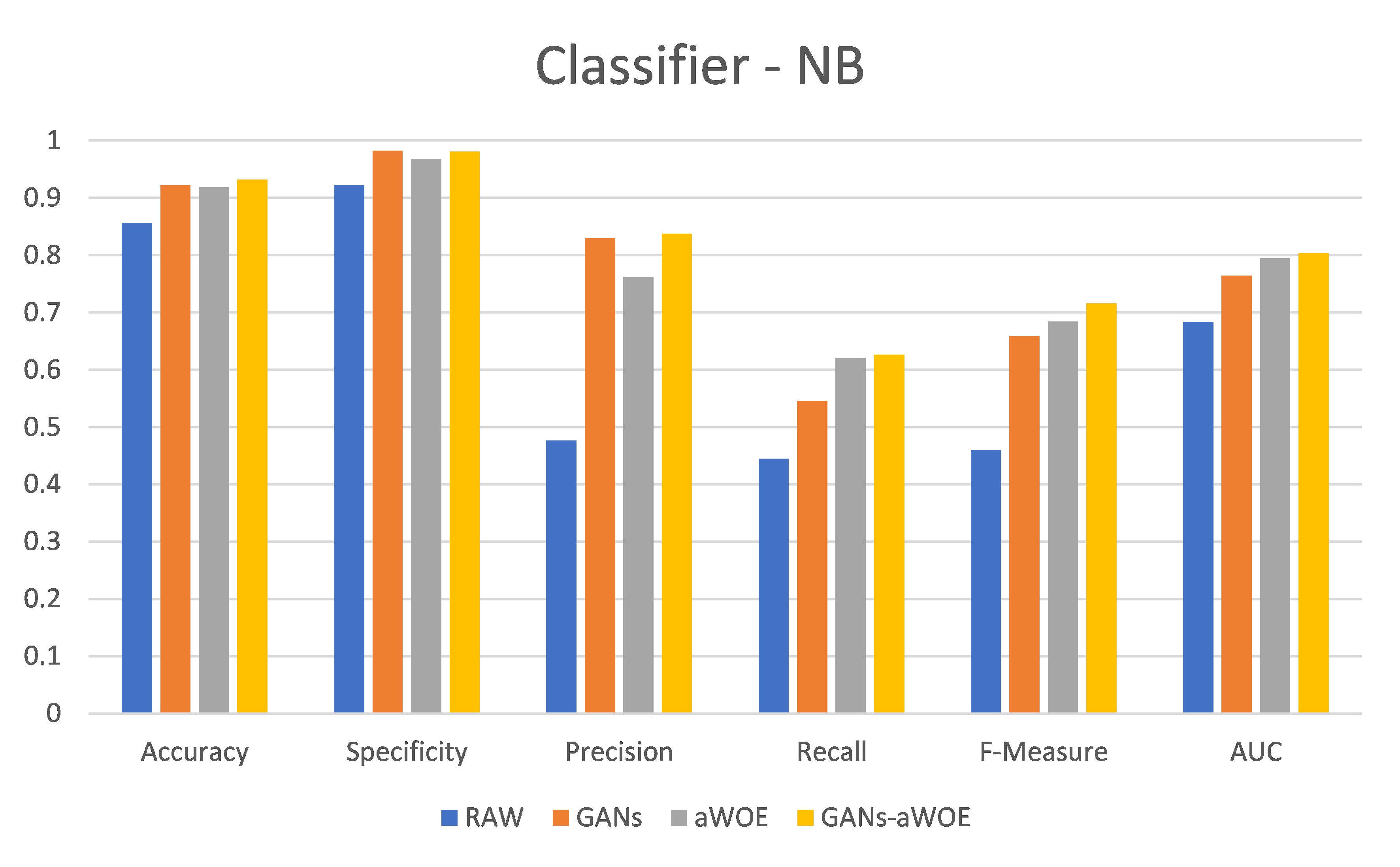}
        \caption{}
    \end{subfigure}
    \begin{subfigure}{0.3\textwidth}
        \includegraphics[width=\linewidth]{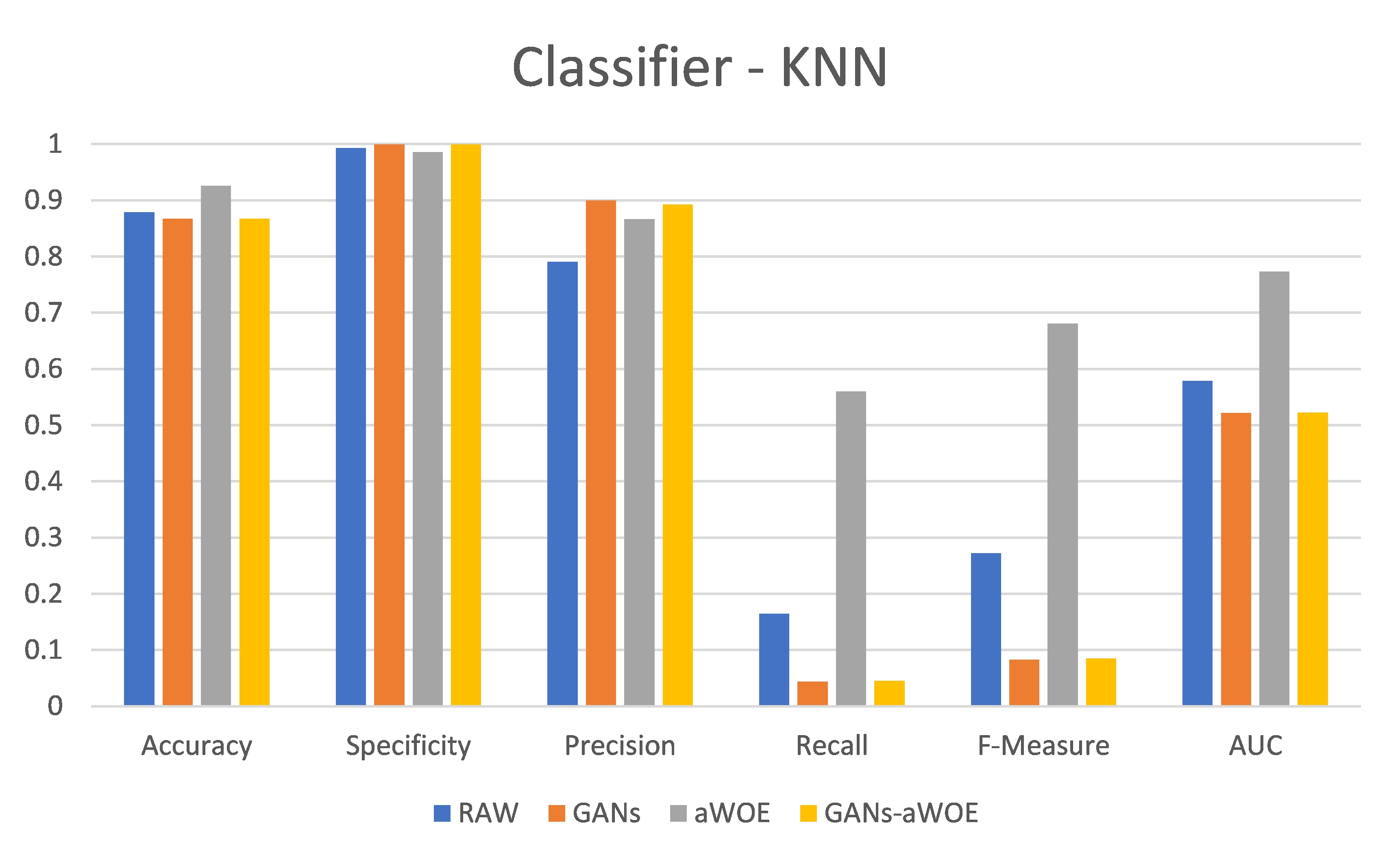}
        \caption{}
    \end{subfigure}
    \begin{subfigure}{0.3\textwidth}
        \includegraphics[width=\linewidth]{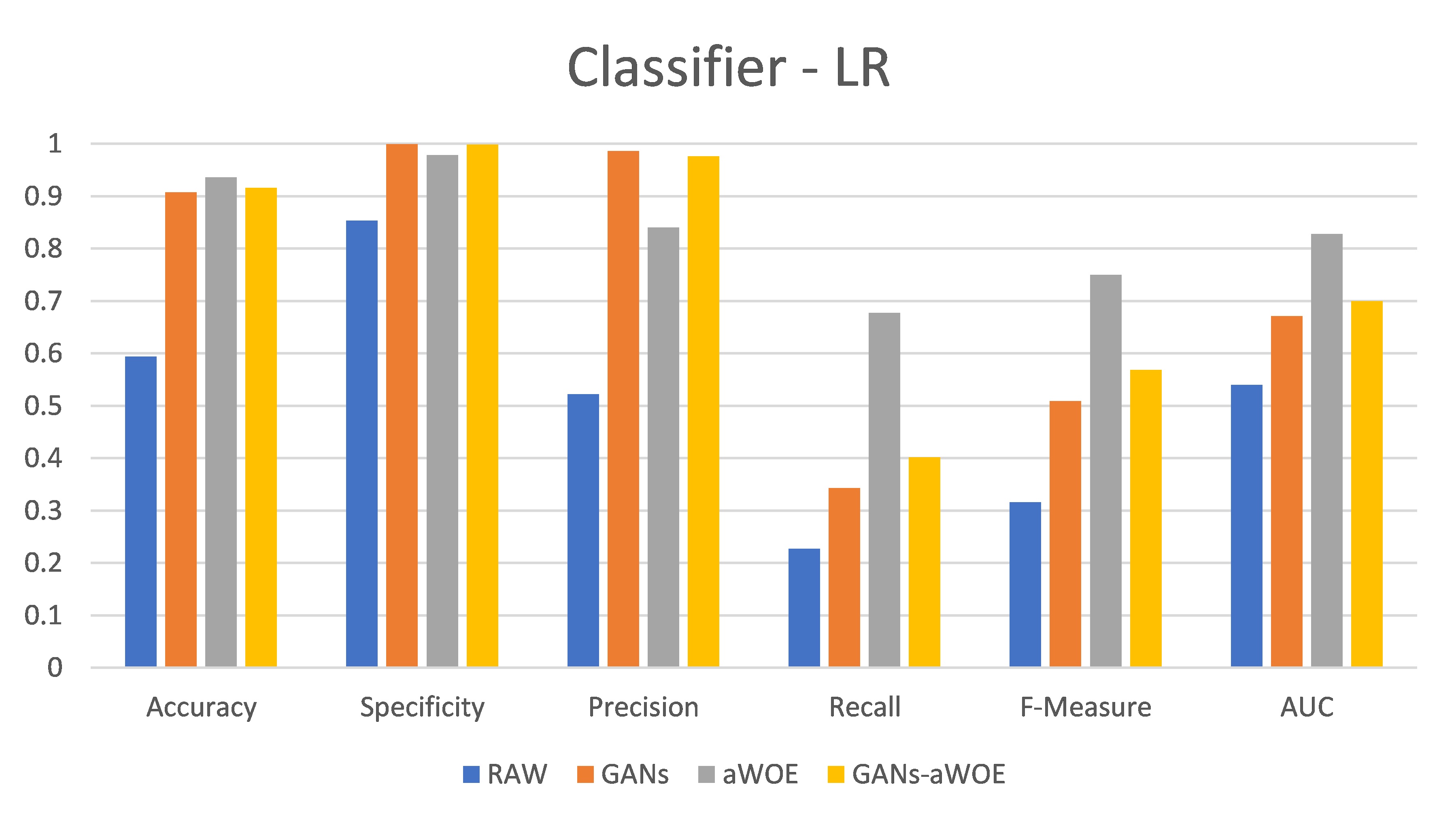}
        \caption{}
    \end{subfigure}
    
    \begin{subfigure}{0.3\textwidth}
        \includegraphics[width=\linewidth]{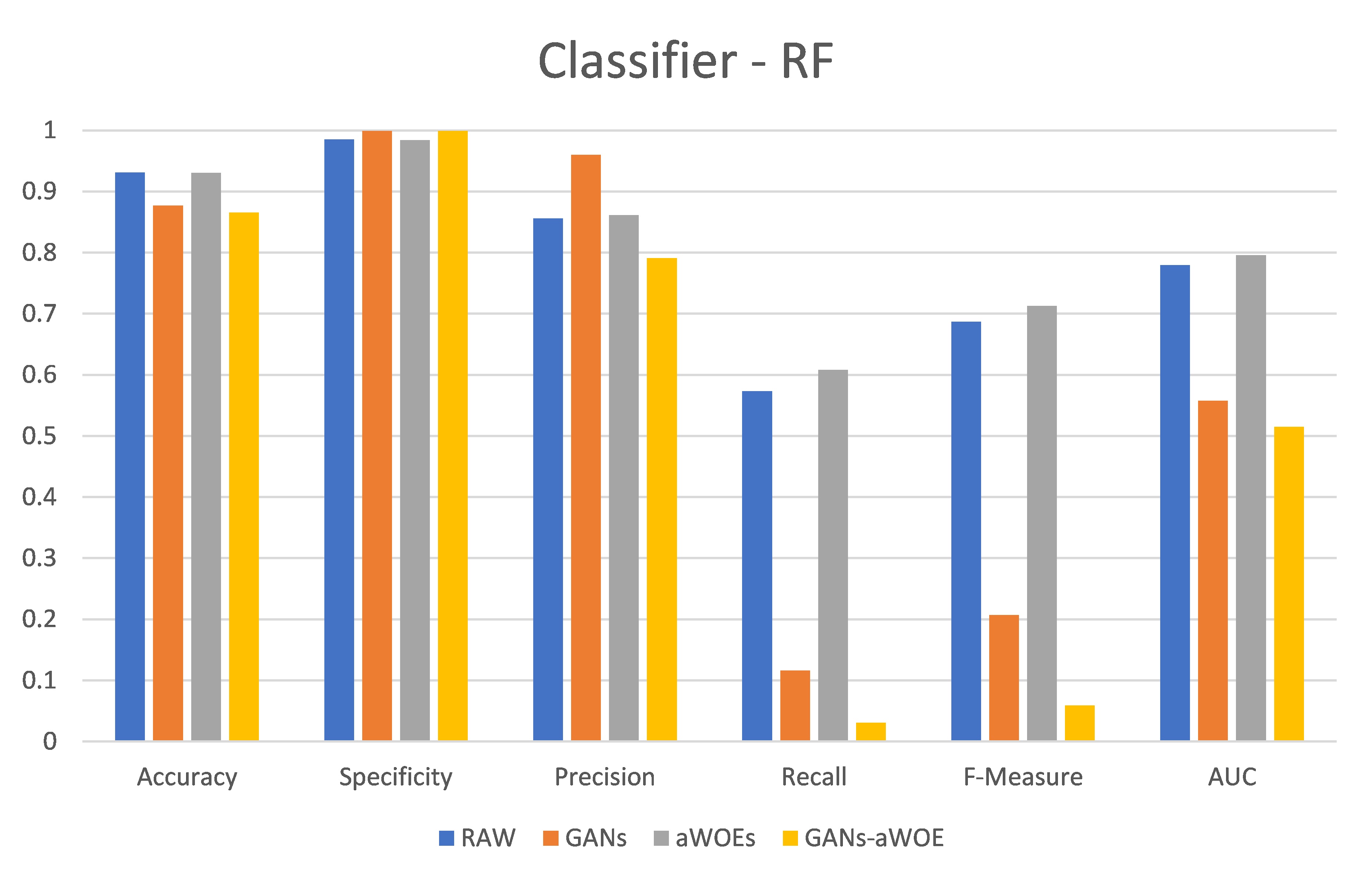}
        \caption{}
    \end{subfigure}
    \begin{subfigure}{0.3\textwidth}
        \includegraphics[width=\linewidth]{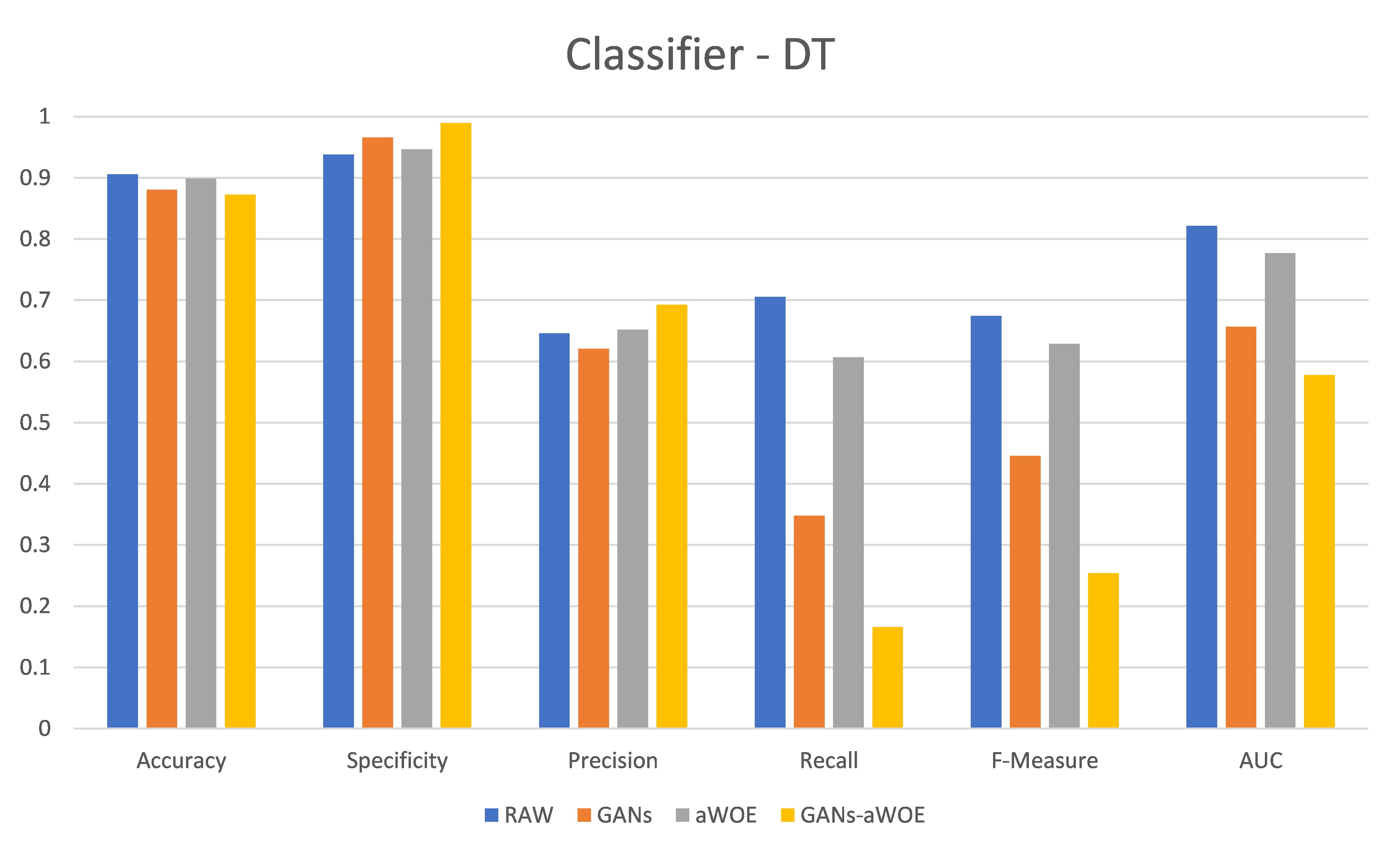}
        \caption{}
    \end{subfigure}
    \begin{subfigure}{0.3\textwidth}
        \includegraphics[width=\linewidth]{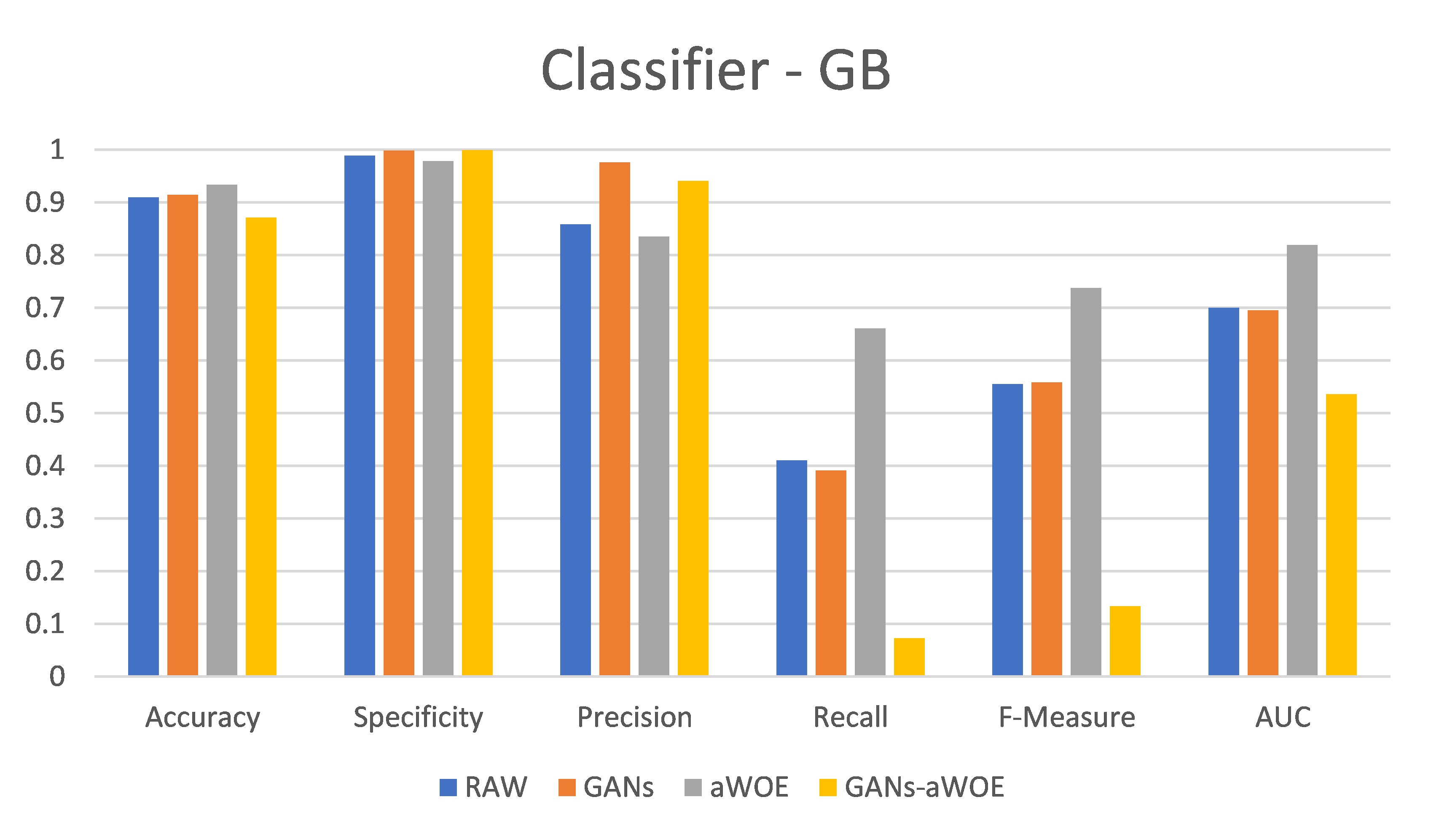}
        \caption{}
    \end{subfigure}

    \begin{subfigure}{0.3\textwidth}
        \includegraphics[width=\linewidth]{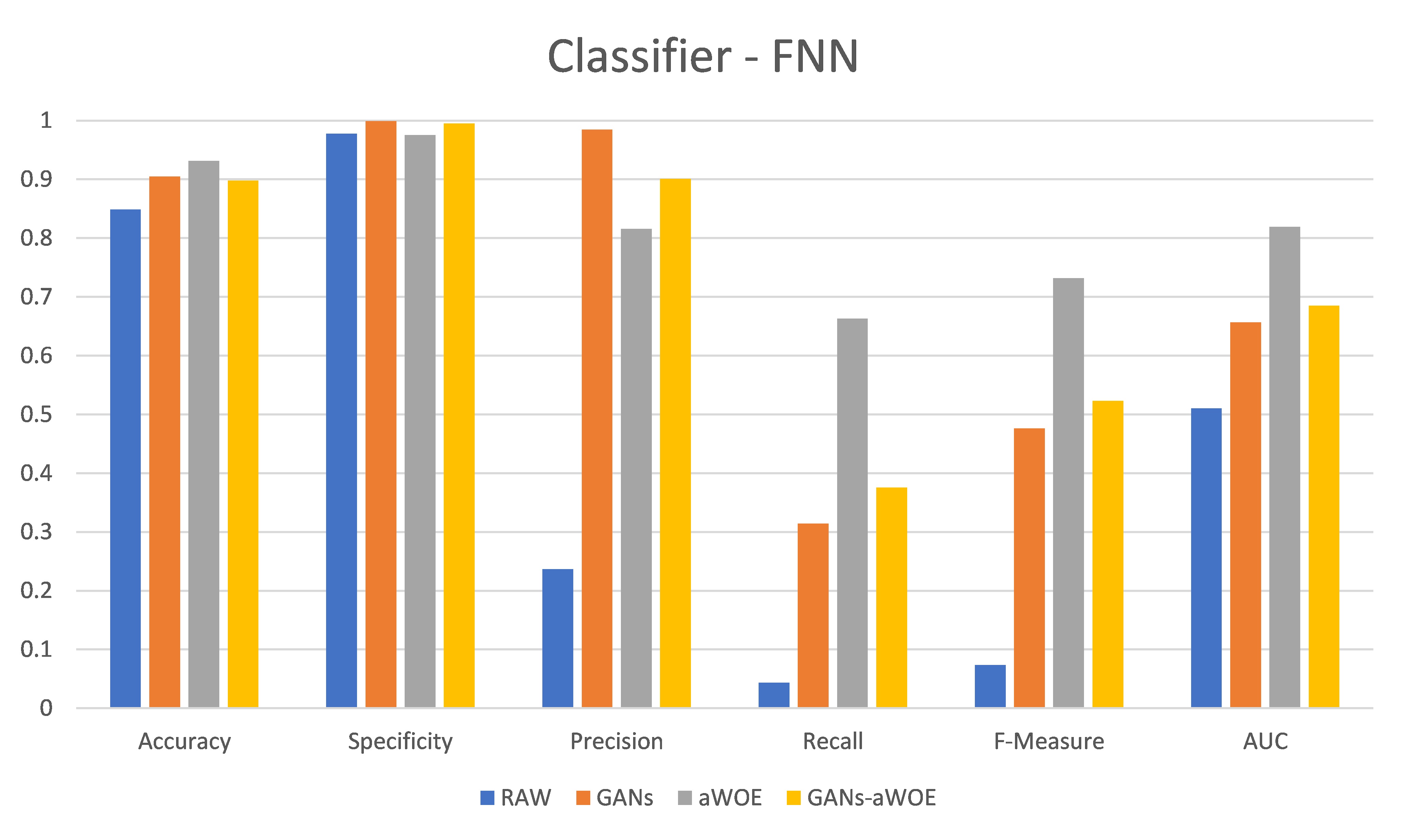}
        \caption{}
    \end{subfigure}
    \begin{subfigure}{0.3\textwidth}
        \includegraphics[width=\linewidth]{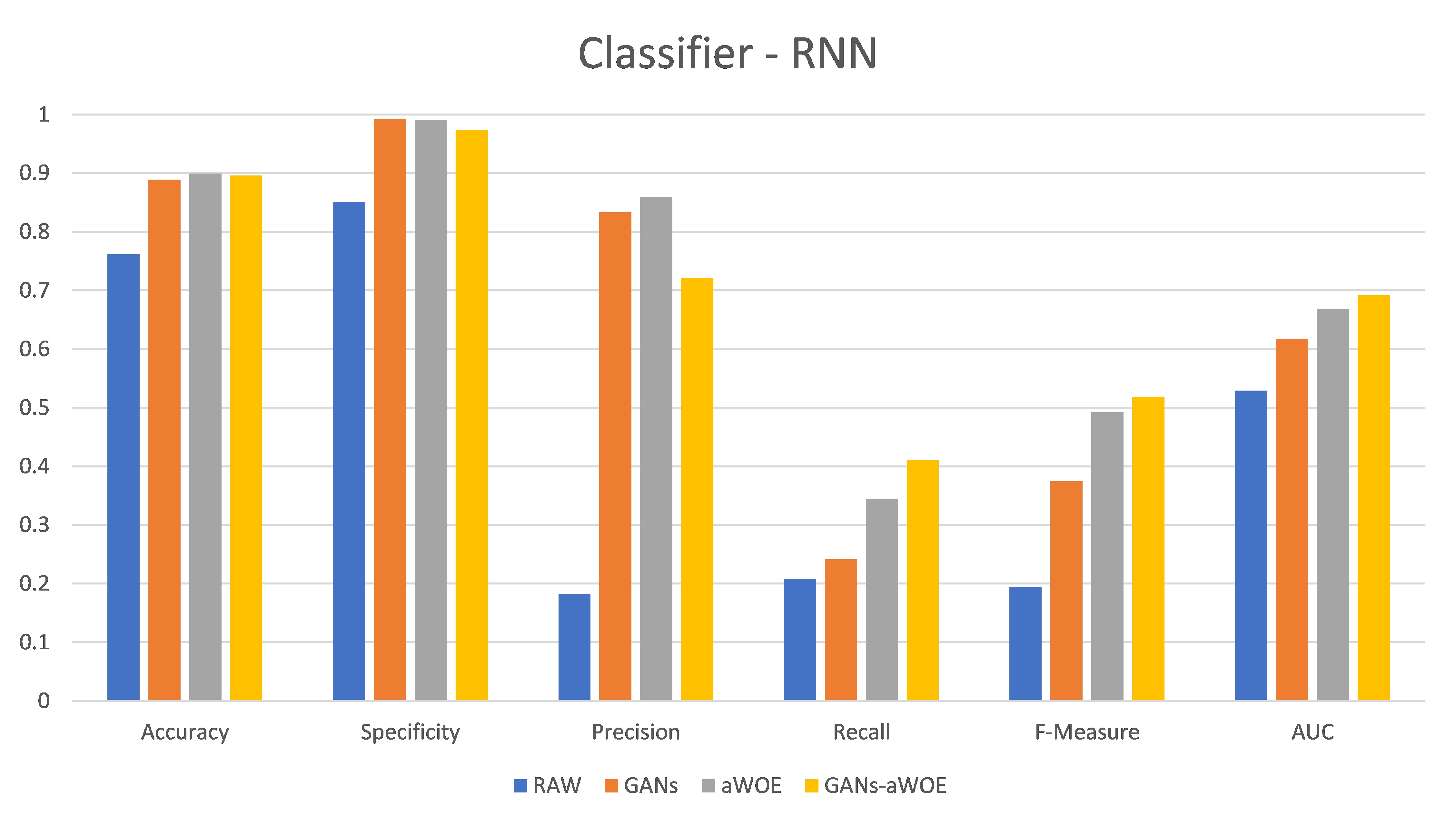}
        \caption{}
    \end{subfigure}
     \caption[]{Performance comparison among the RAW, GANs, aWOE and  GANs-aWOE based CCP models on dataset-3}
  \label{fig:comparison_dataset_3}
\end{figure}

In dataset-3, we noticed similarly impressive results (Figure \ref{fig:comparison_dataset_3}). With a few exceptions, in most cases, the aWOE based classifiers and GANs-aWOE based classifiers displayed superior performance compared to the RAW and GANs based classifiers. For the GANs-aWOE, the best performer in terms of accuracy is NB with a value of 0.932 and in terms of F-measure, the best value achieved by NB is 0.716. In terms of AUC, the highest value achieved by aWOE based FNN which is 0.819. On the other hand, the RAW based NB shows 0.856 and 0.46 in terms of accuracy and F-measure, respectively. 

 \begin{figure}[h!]
    \centering
    \begin{subfigure}{0.3\textwidth}
        \includegraphics[height=135px,width=140px]{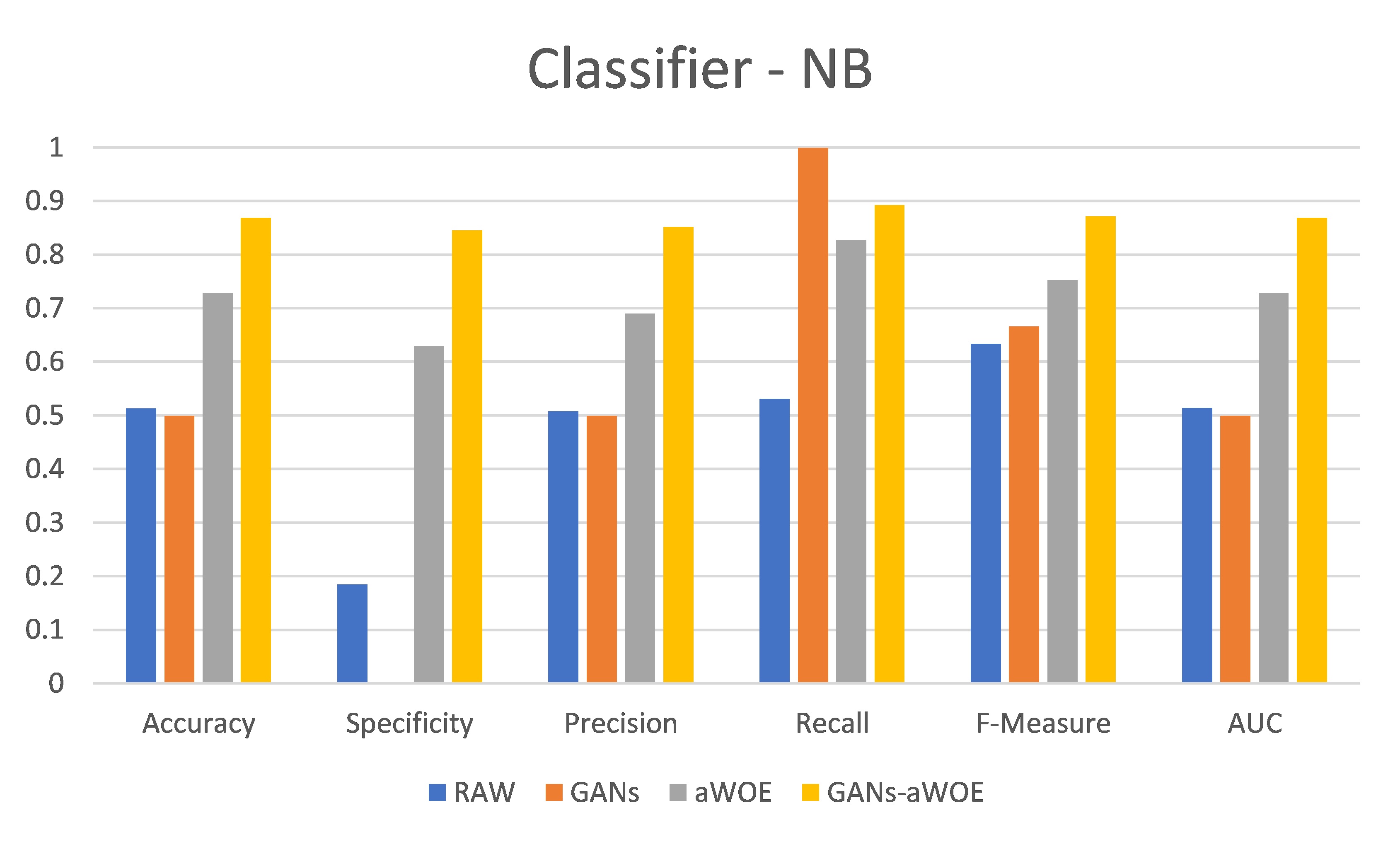}
        \caption{}
    \end{subfigure}
    \begin{subfigure}{0.3\textwidth}
        \includegraphics[height=135px,width=140px]{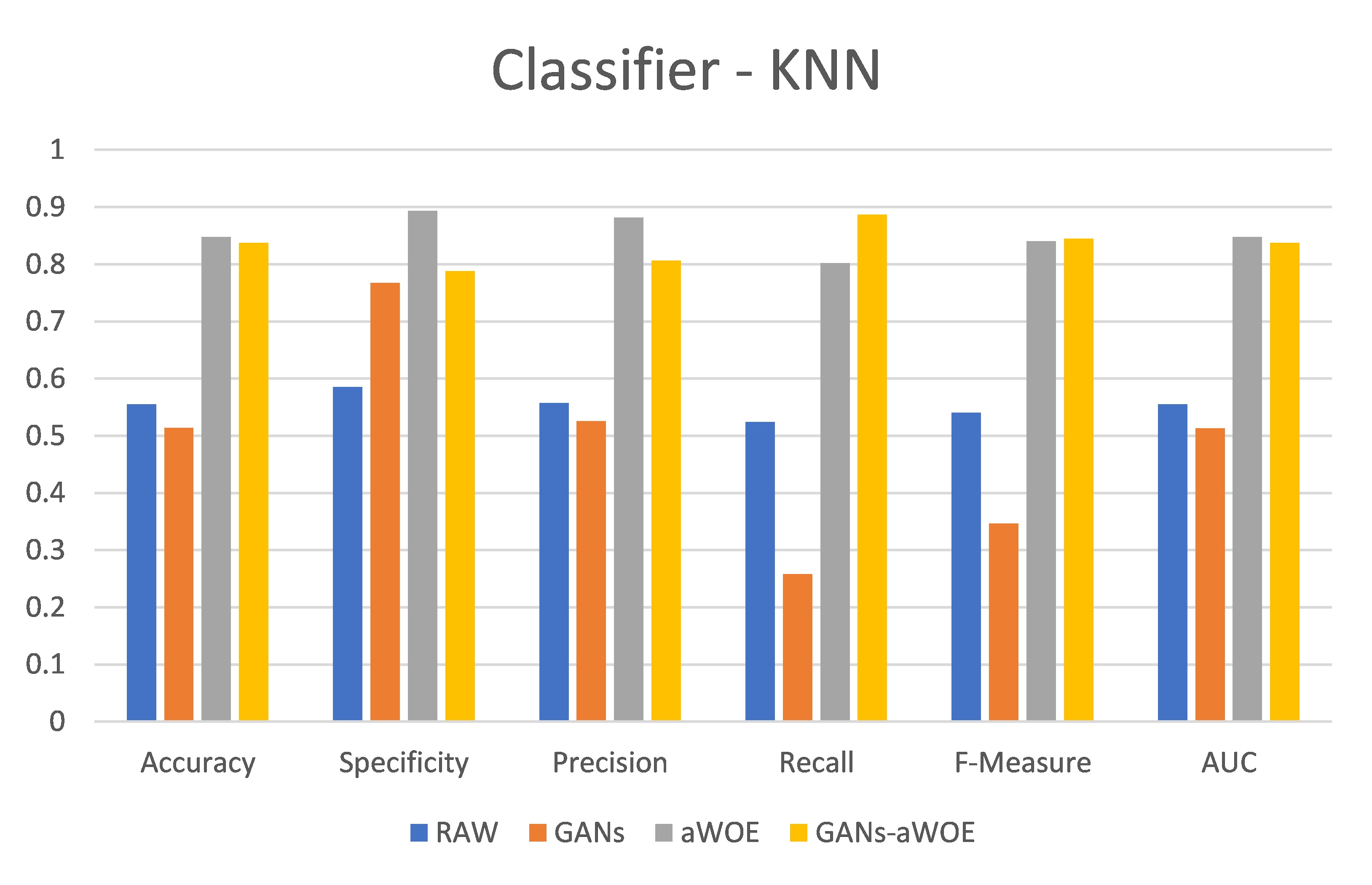}
        \caption{}
    \end{subfigure}
    \begin{subfigure}{0.3\textwidth}
        \includegraphics[height=135px,width=140px]{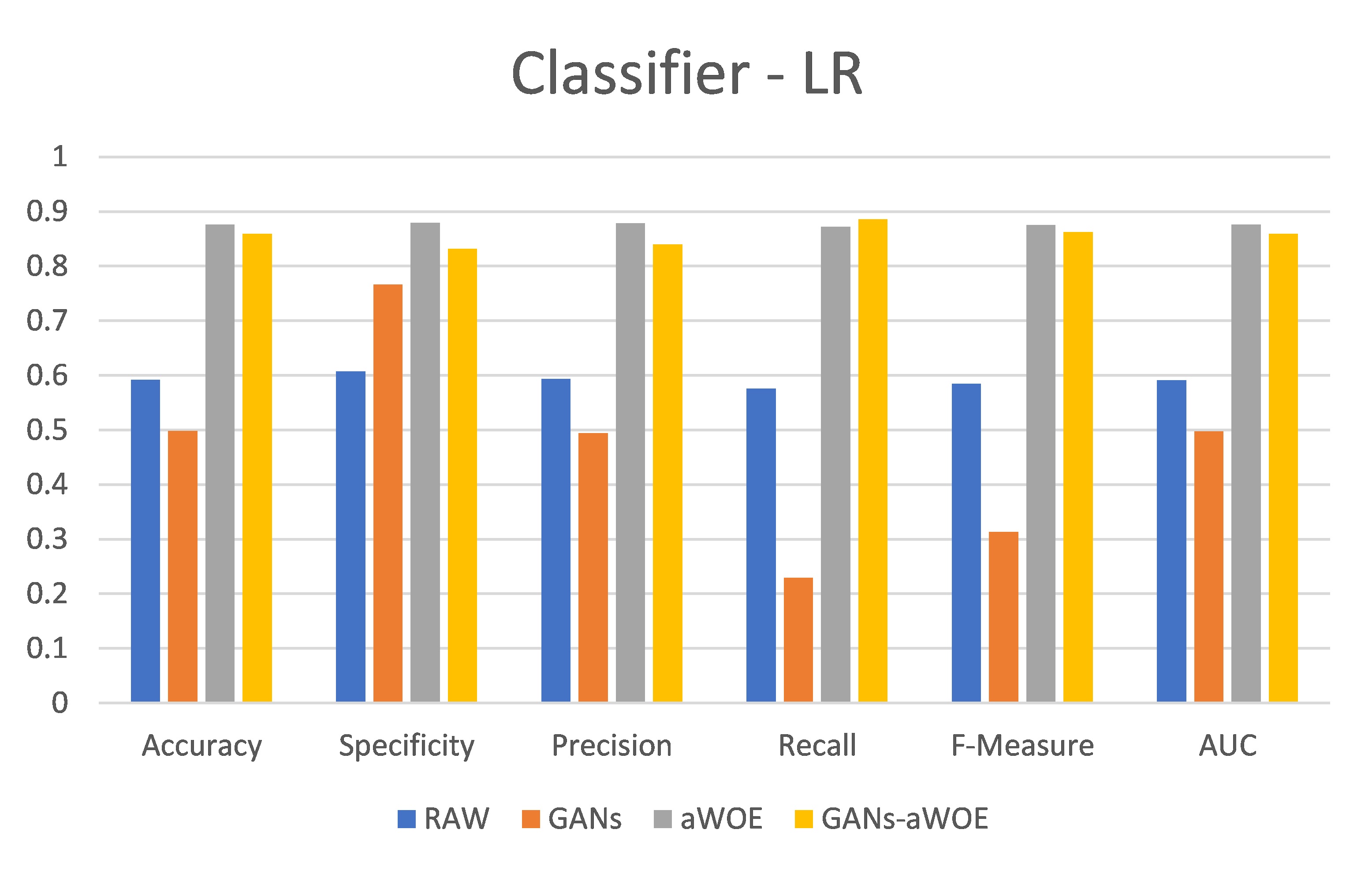}
        \caption{}
    \end{subfigure}
    
    \begin{subfigure}{0.3\textwidth}
        \includegraphics[height=135px,width=140px]{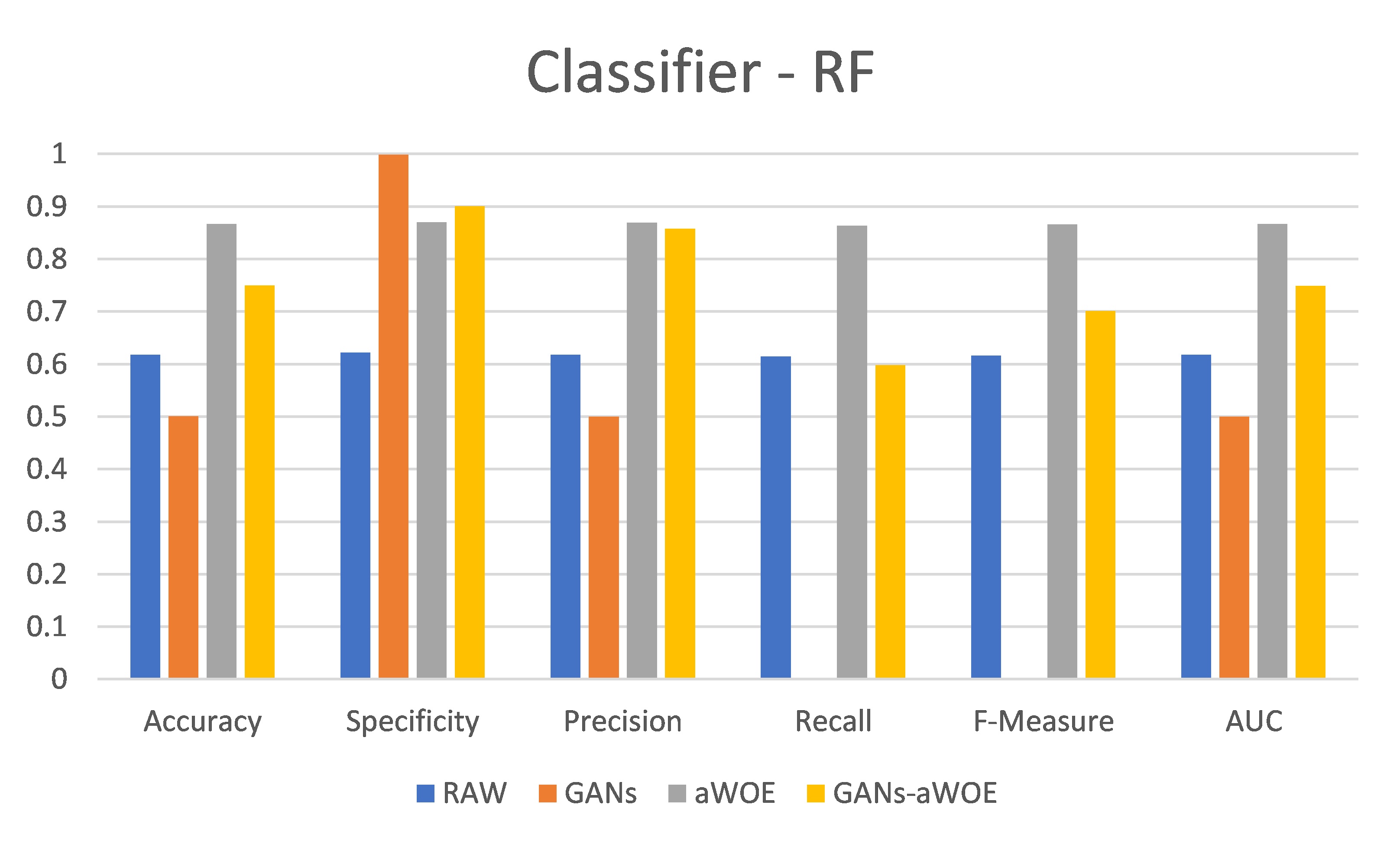}
        \caption{}
    \end{subfigure}
    \begin{subfigure}{0.3\textwidth}
        \includegraphics[height=135px,width=140px]{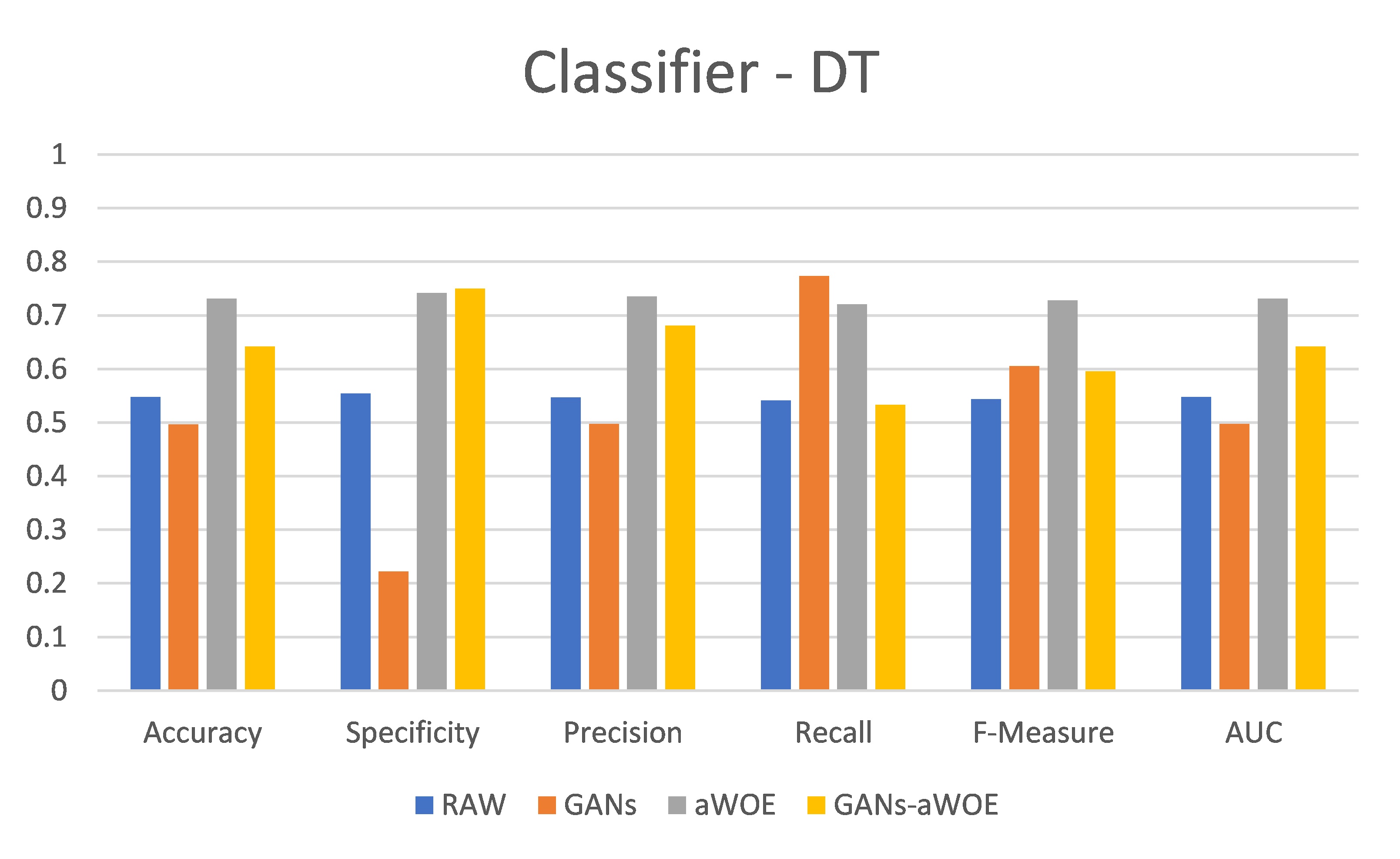}
        \caption{}
    \end{subfigure}
    \begin{subfigure}{0.3\textwidth}
        \includegraphics[height=135px,width=140px]{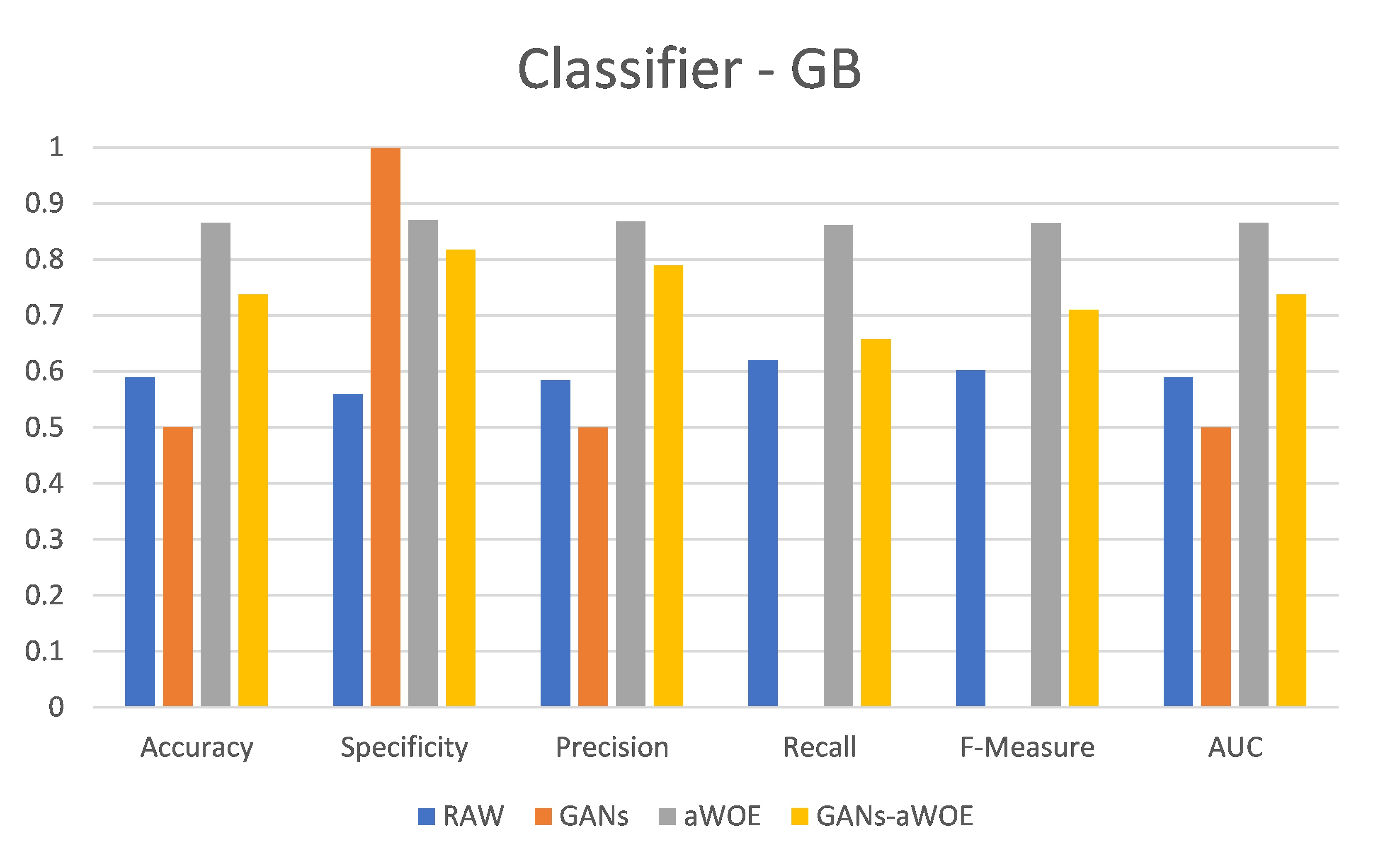}
        \caption{}
    \end{subfigure}
    \begin{subfigure}{0.3\textwidth}
        \includegraphics[height=135px,width=140px]{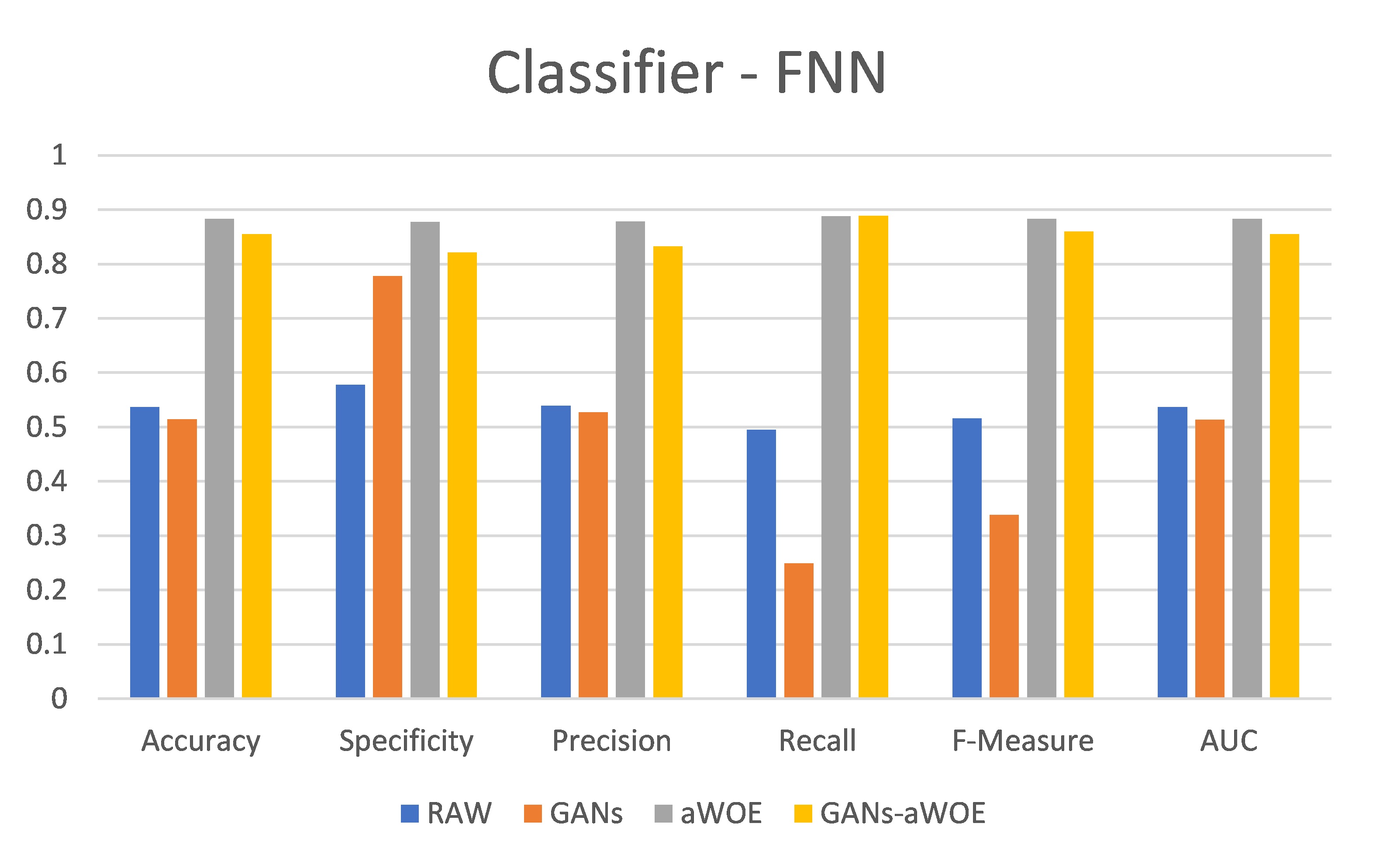}
        \caption{}
    \end{subfigure}
    \begin{subfigure}{0.3\textwidth}
        \includegraphics[height=135px,width=140px]{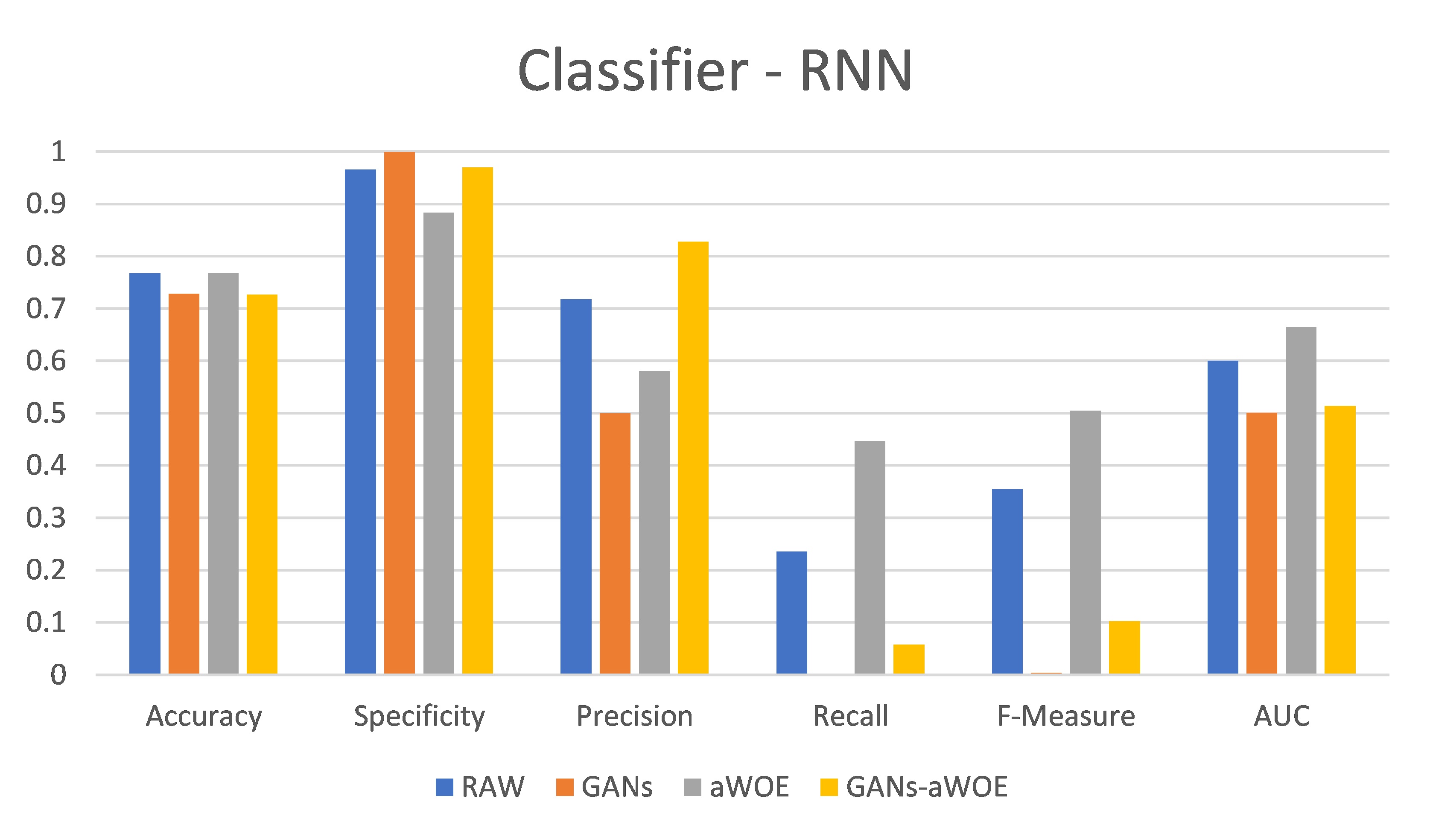}
        \caption{}
    \end{subfigure}
    \caption{Performance comparison among the RAW, GANs, aWOE and  GANs-aWOE based CCP models on dataset-1}
     \label{fig:comparison_dataset_1}
\end{figure}

\subsection{Privacy Analysis}\label{sec:Security_Analysis} 
\begin{figure}[!htb]
\begin{center}
\includegraphics[height=200px,width=250px]{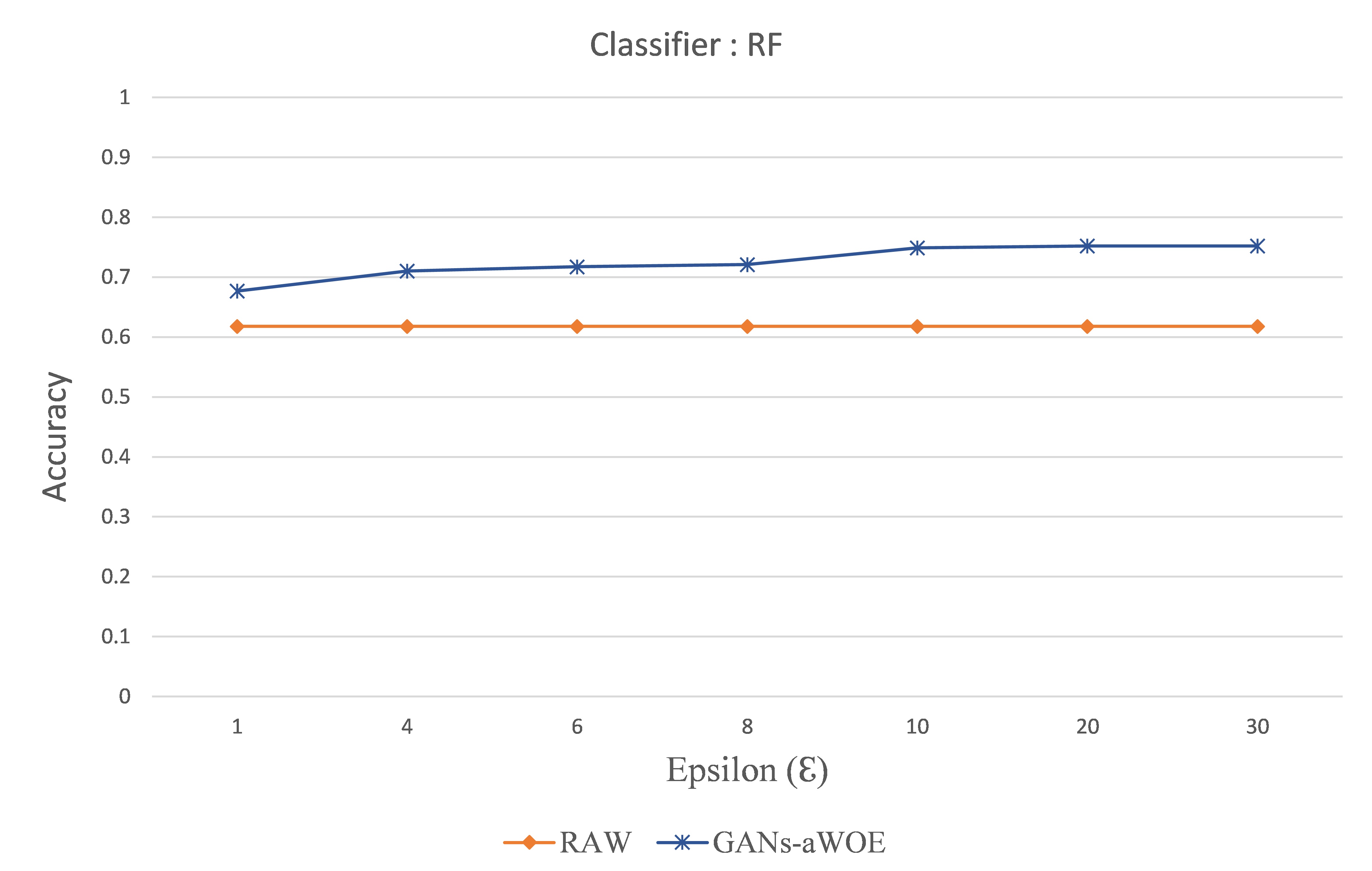}
\caption{The effect of $\epsilon$ on prediction performance in terms of accuracy (RF classifier in dataset-1).}
\label{fig:accuracy-Epcilon-RF-Dataset-1}
\end{center}
\end{figure}

To understand why our processes preserve data privacy, it is important to notice how we process the training data before feeding it to the machine learning classifiers. To generate the synthetic data, we trained the GANs with the privacy parameter $ \epsilon= 10$ (as recommended in \cite{Yang_2020}, \cite{Brett_2018} \cite{xie2018differentially}). We achieve differential privacy by limiting the $\epsilon$ value that controls the maximum influence of any single participant during training and this mechanism adds a certain amount of random noise. This mechanism guarantees the data privacy as discussed in section \ref{subsec:DP} and \ref{subsec:GANs_DP}. Figure \ref{fig:accuracy-Epcilon-RF-Dataset-1} shows the effect of $\epsilon$  on prediction performances in terms of accuracy for the dataset-1.  Lower $\epsilon$ values indicate lower prediction performances but higher data privacy. In our experiments, tree-based classifiers exhibit more observable variations in the performance curve in response to changes of the $\epsilon$ value. Therefore, we illustrate RF classifier based performance curve on dataset-1 in terms of accuracy. Upon obtaining the synthetic data, aWOE is applied on the dataset, which adds another privacy layer on the dataset. aWOE works as a $k$-anonymity ( for our case $k$ changes dynamically) data privacy technique which is proposed by Samarati and Sweeney \cite{Samarati1998}. Suppression and Generalization are the two common techniques used to achieve $k$-anonymity. Generalization based $k$-anonymity technique has been used in the aWOE data transformation approach. For a particular bin, all values are replaced with the same aWOE value. In general, on average each bin contains n/b number of samples (n is the total number of training samples). When the number of training records n will be bigger, the average data size (n/b) of a bin will be bigger and the data value will be replaced with the same aWOE value, which makes it very difficult for the cloud server (or an adversary) to retrieve individual records from the training dataset. It is also difficult for an adversary to guess a real individual record from the output of the trained model as the model has been trained using synthetic data which is produced based on differential privacy and aWOE based transformation is applied atop it. Differential privacy is a tool to serve as an effective safeguard for privacy against membership inference attacks \cite{PAN_2024, Chen2021}. DP also provides a protective shield against model inversion attack \cite{PAN_2024}. As our training datasets are synthesized with differential privacy. our trained models are protected from those attacks.

\subsection{Statistical Performance Analysis}\label{sec:Statistical_Analysis} 
In this study, we consider Wilcoxon signed-rank test~\cite{Janez2006} to compare the performance of the various classifiers to determine whether the proposed GANs-aWOE based classifiers and the RAW based classifiers perform equally or there are significant differences in their performance. The statistical test has been performed in terms of accuracy and F-measure at 0.05 level of significance $(\alpha)$. The Wilcoxon signed-rank non-parametric test was chosen because it does not require any underlying data distribution. Here the the null hypothesis $(H_0)$ is there is no significant difference between the RAW based classifies performances and GANs-aWOE based classifies performances. Table \ref{table:WillcoxsonTest} presents the Z-values and corresponding p-values of the Wilcoxon signed-rank test result. For both accuracy and F-measure, the Wilcoxon signed-rank test rejects the null hypothesis $(H_0)$. Based on the Wilcoxon signed-rank test result and Figure \ref{fig:comparison_dataset_1}, \ref{fig:comparison_dataset_2}, \ref{fig:comparison_dataset_3},  it is evident that the performance of GANs-aWOE based models significantly outperform the RAW based models.

To determine which classifier is performing better among the GANs-aWOE based classifiers, we calculated the average rank of the classifiers. We rank the models for each data set separately, the best performing model gets the rank of 1, the second best rank 2 and so on. Then the average rank has been calculated using the following formula.
 \begin{equation} \label{eq:average_rank}
     R_j = \frac{1}{N}\sum_{i}{r}_i^j
\end{equation}

 Where ${r}_i^j$ is the rank of the $j$-th of $k$ models on the $i$-th of N data sets. Table \ref{table:AVG_RANK_accuracy} and Table \ref{table:AVG_RANK_F_measure} summarize the ranking of the GANs-aWOE based classifiers performances across all datasets. Ranking tables show that the NB classifier is the best performer, followed by LR.

\begin{table}[hbt!] 
\begin{center}
\caption{ Wilcoxon signed-rank test result }
\label{table:WillcoxsonTest}
{
\renewcommand{\arraystretch}{1.5}
\begin{tabular}{ p{5cm} p{2cm} p{2cm}  p{2cm}  p{2cm} }
\hline
Measurement Metric & z-value & p-value & Hypothesis ($\alpha=0.05$) \\ \hline
Accuracy & -3.3714 &0.00076&Rejected\\
F-measure &-2.0286 &0.04236&Rejected\\
\hline
 \vspace{.01mm}\\
\end{tabular}
}
\end{center}
\end{table}

\begin{table}[h!]
 \centering
 \caption{Average rank of the classifiers for GANs-aWOE method across all datasets in terms of accuracy.}
\label{table:AVG_RANK_accuracy}
{
\renewcommand{\arraystretch}{1.5}
\begin{tabular}{p{5cm} p{5cm} }\hline
\textbf{Classifier } & \textbf{Average Rank }\\
\hline

NB & 1\\
LR & 2 \\
FNN & 3 \\
KNN & 5 \\
RNN & 6 \\
RF & 6 \\
GB & 6.3333 \\
DT & 6.6667 \\
\hline
\end{tabular}
}
\end{table}

\begin{table}[h!]
 \centering
 \caption{Average rank of the classifiers for GANs-aWOE method across all datasets in terms of F-measure. }
\label{table:AVG_RANK_F_measure}
{
\renewcommand{\arraystretch}{1.5}
\begin{tabular}{p{5cm} p{5cm} }\hline
\textbf{Classifier } & \textbf{Average Rank }\\
\hline

NB & 1.6667 \\
LR & 1.6667 \\
FNN & 2.6667 \\
KNN & 4.6667 \\
RNN & 5.6667 \\
RF & 6.3333 \\
GB & 6.3333 \\
DT & 7 \\
\hline
\end{tabular}
}
\end{table}

 \subsection{Comparison with other studies}\label{sec:comparison_other_studies}
 In this study, we focused on data privacy based customer churn prediction for the telecommunication industry. To the best of our knowledge, data privacy for the customer churn prediction in TCI has not been studied yet. However, it's important to mention that all three datasets used in this study have also been utilized in a handful of earlier research studies like sana et al.~\cite{sana_JK_2022}, and Amin et al.~\cite{AMIN_2019_BR}. But they refer to these datasets with slightly different names. Hence a mapping between the dataset names is outlined in Table \ref{table:Dataset_mapping_with_previous_study}. Tables \ref{table:PER_COMP_pre_study_Dataset_1}, \ref{table:PER_COMP_pre_study_Dataset_2}, and \ref{table:PER_COMP_pre_study_Dataset_3} illustrate the performance comparison among the proposed best two GANs-aWOE based models and previously mentioned studies (~\cite{sana_JK_2022}, ~\cite{AMIN_2019_BR}), in terms of accuracy, F-measure, precision, recall, and AUC. Those tables show proposed models achieve better performance and improve prediction performance up to 27.9\%  and 6\% in terms of F-measure for the dataset 1 and 2, respectively. For the dataset-3 our proposed models show slightly higher performance than the previous study ~\cite{sana_JK_2022} in term of accuracy and precision but in term of F-measure and AUC, it is slightly lower. This is likely due to the small sample size of the dataset and the associated imbalance.

 \begin{table}[hbt!] 
\begin{center}
\caption{Mapping of dataset names in this study and previous studies in \cite{sana_JK_2022} and \cite{AMIN_2019_BR}}
\label{table:Dataset_mapping_with_previous_study}
{ 
\renewcommand{\arraystretch}{1.5}
\begin{tabular} { p{2.5 cm} p{2.5  cm} p{2.5 cm} }
\hline
 This Study & Ref. Study ~\cite{sana_JK_2022} & Ref. Study ~\cite{AMIN_2019_BR}\\ 
\hline
  Dataset-1 &  Dataset-1 & Dataset-4 \\
  Dataset-2 &  Dataset-4 & Dataset-2 \\
  Dataset-3 &  Dataset-2 & - \\
\hline
 \vspace{.01mm}\\
 
\end{tabular}
}
\end{center}
\end{table}

 \begin{table}[hbt!] 
\begin{center}
\caption{ Performance comparison between this study and previous studies in \cite{sana_JK_2022} and \cite{AMIN_2019_BR}  using Dataset-1}
\label{table:PER_COMP_pre_study_Dataset_1}
{ 
\renewcommand{\arraystretch}{1.5}
\begin{tabular} { p{4 cm} p{2.0cm} p{1.75cm} p{1.75cm} p{1.5cm} p{1.5cm} p{1.5cm}}
\hline
Ref. Study & Data Privacy & Accuracy & F-Measure & Precision & Recall & AUC\\ 
\hline
 \centering GANs-aWOE based NB model (current study) & \centering YES & \textbf{0.869} & \textbf{0.871} & \textbf{0.851} & \textbf{0.892} & \textbf{0.869}\\
 \centering GANs-aWOE based LR model (current study) & \centering YES & 0.859 & 0.869 & 0.82 & 0.886 & 0.860\\
 \centering Ref. \cite{sana_JK_2022}  & \centering NO & 0.802 & 0.80 & 0.805 & 0.802 &0.802\\
\centering Ref. \cite{AMIN_2019_BR} & \centering NO & 0.58 & 0.592 & 0.567& 0.626 &- \\
\hline
 \vspace{.01mm}\\
 
\end{tabular}
}
\end{center}
\end{table}

\begin{table}[hbt!] 
\begin{center}
\caption{ Performance comparison between this study and previous studies in \cite{sana_JK_2022} and \cite{AMIN_2019_BR}  using Dataset-2}
\label{table:PER_COMP_pre_study_Dataset_2}
{ 
\renewcommand{\arraystretch}{1.5}
\begin{tabular}{ p{4 cm} p{2.0cm} p{1.75cm} p{1.75cm} p{1.5cm} p{1.5cm} p{1.5cm}}
\hline
  Ref. Study &  Data Privacy & Accuracy & F-Measure & Precision & Recall & AUC\\ 
\hline
 \centering GANs-aWOE based NB model (current study) & \centering YES & \textbf{0.832} & 0.671 & \textbf{0.718} & 0.632 & 0.770\\
 \centering GANs-aWOE based LR model (current study) & \centering YES & 0.828 & \textbf{0.686} & 0.682 & \textbf{0.691} & \textbf{0.785}\\
 \centering Ref. \cite{sana_JK_2022}  & \centering NO & 0.819 & 0.634 & 0.688 & 0.587 &0.746\\
 \centering Ref. \cite{AMIN_2019_BR} & \centering NO & 0.748 & 0.626 & 0.504& 0.824 &- \\
\hline
 \vspace{.01mm}\\
\end{tabular}
}
\end{center}
\end{table}

\begin{table}[hbt!] 
\begin{center}
\caption{ Performance comparison between this study and previous studies in \cite{sana_JK_2022} and \cite{AMIN_2019_BR}  using Dataset-3}
\label{table:PER_COMP_pre_study_Dataset_3}
{ 
\renewcommand{\arraystretch}{1.5}
\begin{tabular}{ p{4 cm} p{2.0cm} p{1.75cm} p{1.75cm} p{1.5cm} p{1.5cm} p{1.5cm}}
\hline
 Ref. Study & Data Privacy & Accuracy & F-Measure & Precision & Recall & AUC\\ 
\hline
 \centering GANs-aWOE based NB model (current study) & \centering YES & \textbf{0.932} & 0.716 & 0.837 & 0.626 & 0.803\\
 \centering GANs-aWOE based LR model (current study) & \centering YES & 0.916 & 0.569 & \textbf{0.976} & 0.402 & 0.700\\
 \centering Ref. \cite{sana_JK_2022}  & \centering NO & 0.931 & \textbf{0.732} & 0.816 & \textbf{0.663} & \textbf{0.819}\\
\centering Ref. \cite{AMIN_2019_BR} & \centering NO & - & - & - & - &- \\
\hline
 \vspace{.01mm}\\
\end{tabular}
}
\end{center}
\end{table}
 
\subsection{Impact of GANs and aWOE on prediction performance}\label{sec:Why_improves} 
The synthetic datasets are generated from GANs models. As the GANs are trained within differential privacy budget, we expect that it changes the data distribution slightly which is shown in study \cite{xie2018differentially}. However, utilization of aWOE on the GANs generated synthetic data provides less skewed dataset, as illustrated in Table \ref{table:skew_value_Dataset_3} (lower absolute skew value represents less skewed data). Research in \cite{sana_JK_2022} and \cite{Amin_Adnan_2019} show that less skewed data improves the prediction performance. Therefore, we believe that this property of our approach helps to enhance the prediction performance in all cases.

\begin{table}[]
\begin{center}
\caption{ Skew value of first 10 features of dataset-1 for the GANs and GANs-aWOE approaches}
\label{table:skew_value_Dataset_3}
{ 
\renewcommand{\arraystretch}{1.5}
\begin{tabular}{ p{4cm} p{4cm} p{4cm}}
\hline
\centering \textbf{Features Name}            & \textbf{GANs}         & \textbf{GANs-aWOE} \\ \hline
\centering  rev\_Mean        & -0.308922507 & 0.041761  \\  
\centering  mou\_Mean        & -0.089470317 & 0.005585  \\   
\centering  totmrc\_Mean      & -0.225770289 & 0.061772  \\  
\centering  da\_Mean         & -0.085141633 & 0.04489   \\  
\centering  ovrmou\_Mean     & -0.276226956 & 0.016191  \\  
\centering  ovrrev\_Mean     & -0.15172743  & 0.062511  \\  
\centering  vceovr\_Mean      & 0.259959698  & -0.02976  \\  
\centering  datovr\_Mean     & -0.216810421 & -0.01569  \\  
\centering  roam\_Mean       & -0.157843443 & -0.00501  \\  
\centering  drop\_vce\_Mean  & -0.237205413 & -0.02944  \\ \hline
\end{tabular}
}
\end{center}
\end{table}

\section{Discussion}  \label{sec:Discussion} 
This study is mostly designed for privacy preserving CCP model for the telecommunication industry. To preserve the data privacy, we generate training data from original RAW data using GANs models within the differential privacy budget limit. The privacy analysis (section \ref{sec:Security_Analysis}) manifests that the GANs-aWOE technique allows the telecom industry to train their model in third-party cloud servers without compromising the data privacy. Trained machine learning model using this privacy protected data also protect privacy from various adversarial attacks and data leakage. The GANs-aWOE model allows the data owner to share the data with other parties while addressing legal and ethical considerations.

In this research work, we proposed an adaptive Weight-of-evidence (aWOE) data transformation method which enhances the prediction performance greatly as discussed in the section \ref{sec:aWOE_WOE}. We used this aWOE technique for our proposed privacy preserving CCP model. To examine whether our proposed technique achieves better prediction performance, six widely used measurement metrics (i.e. accuracy, specificity, precision, recall, F-measure, and AUC) have been used to assess the models. In this study, we considered eight different state-of-the-art machine learning classifiers combined with our proposed GANs-aWOE technique on three publicly available datasets. Our experimental results and comparative analysis involving the RAW based models and GANs-aWOE models clearly indicate that GANs-aWOE technique improves the prediction performance by a wide margin (up to 35.6\% in terms of accuracy) (Figure \ref{fig:comparison_dataset_1} - \ref{fig:comparison_dataset_3} and supplementary Tables \ref{table:GANs_WOE_methods_on_dataset-1}, \ref{table:GANs_WOE_methods_on_dataset-2} and \ref{table:GANs_WOE_methods_on_dataset-3}). The outcome of this research can provide a significant advantage in the existing and future prediction models for telecom companies. The GANs-aWOE based approach provides uniform performance improvement as observed across all our experiments except for the RNN for dataset-2 and tree based classifiers (RF, DT, and GB) for datasets-3, in terms of F-measure. We suspect that due to the imbalanced and small size of those two datasets, the binning technique in the aWOE on the GANs generated data is not suitable enough for those classifiers.  It is expected that RNN is not suitable for no-sequential dataset which is also shown true in our study. However, we used the RNN in because a previous study \cite{Chopra_2017} claimed that RNN works fine in their case (no-sequential data). The statistical test shows that GANs-aWOE based models achieve significantly better performance than RAW based models. The statistical analysis also shows that NB is the best classifier and the rank values are 1 and 1.6667 in terms of accuracy and F-measure, respectively (From the table \ref{table:AVG_RANK_accuracy} and \ref{table:AVG_RANK_F_measure}). The second highest rank was achieved by the LR classifier and the rank values are 2 and 1.6667 in terms of accuracy and F-measure, respectively.

To the best of our knowledge, this is the first study of privacy preserving CCP model in the telecom industry. Comparison with previous studies (Section \ref{sec:comparison_other_studies}) shows that the proposed technique achieves better performance than the previous studies and the performance improves up to 28.9\% and 27.9\% in terms of accuracy and F-measure, respectively. We believe that using this technique data owner can share their data to others for various purposes while ensuring data privacy and maintaining data usability. Researchers can utilize this methodology for further research perspectives, while engineers in the telecom industry can leverage it for developing successful data privacy-based models.

\section{Conclusion}  \label{sec:Conclusions} 
Data privacy of a trained machine learning model and during the training time has been a paramount concern. Challenges and threats in this realm have persisted  ever since the machine learning technology started to evolve. Preserving the privacy of sensitive data stored on a cloud server and conducting model training with this data in the cloud server have become more serious concerns in order to satisfy legal and ethical issues. In this study, we introduce GANs-aWOE based approach to protect the privacy of individual records and to improve the churn prediction performance in the context of the telecom industry when they use cloud service to train a model. In the GANs-aWOE approach, synthetic data has been generated using GANs where differential privacy has been imposed. To increase the degree of data privacy and improve the prediction performance, we implemented aWOE transformation on the synthetic data before uploading the data on the cloud server and feeding the data to the machine learning classifiers. The proposed GNAs-aWOE technique not only protects privacy from the curious cloud service provider and the adversary but also improves the churn prediction performance. We used eight different state-of-the-art machine learning classifiers to test the effectiveness of the proposed technique. We also used both balanced and imbalanced datasets to test the sturdiness of our proposed approach. The performance of the resulting models was evaluated with six widely used evaluation measures which gave consistent and reliable results. The experimental outcomes and performance comparisons indicate the supremacy of the GAN-aWOE approach in terms of churn prediction performance. CCP with Privacy is still remains a formidable challenge for the competitive businesses, especially within the telecommunications sector, as it continues to evolve. Furthermore, there is room for extending our proposed methodology to other sectors, thereby investigating the universality of our assertion across various business domains considering the correlation between differential privacy and model robustness.




\bibliographystyle{unsrt}  

\bibliography{Bibliography}

 \clearpage 

 \appendix

\begin{table}[]
\begin{center}
\caption{RAW, GANs, aWOE and GANs-aWOE based models on Dataset-1. For each performance metric, the best result is shown in \textbf{bold-face}.}
\label{table:GANs_WOE_methods_on_dataset-1}
{\scriptsize
\begin{tabular}{|l|l|l|l|l|l|l|l|}
\hline
Classifier            & Method         & Accuracy & Specificity & Precision & Recall & F-Measure & AUC   \\ \hline
                      & RAW            & 0.513    & 0.185       & 0.507     & 0.531  & 0.634     & 0.514 \\ \cline{2-8} 
                      & GAN            & 0.499    & 0.001       & 0.499     & 0.999  & 0.666     & 0.499 \\ \cline{2-8} 
                      & aWOE           & 0.729    & 0.63        & 0.69      & 0.828  & 0.753     & 0.729 \\ \cline{2-8} 
\multirow{-4}{*}{NB}  & GAN-aWOE (avg) & \textbf{0.869}    & 0.845       & 0.851     & 0.892  & 0.871     & 0.869 \\ \hline
                      & RAW            & 0.555    & 0.586       & 0.558     & 0.525  & 0.541     & 0.555 \\ \cline{2-8} 
                      & GAN            & 0.514    & 0.768       & 0.526     & 0.258  & 0.346     & 0.513 \\ \cline{2-8} 
                      & aWOE           & 0.848    & 0.893       & 0.882     & 0.802  & 0.84      & 0.848 \\ \cline{2-8} 
\multirow{-4}{*}{KNN} & GAN-aWOE (avg) & 0.837    & 0.788       & 0.807     & 0.887  & 0.845     & 0.838 \\ \hline
                      & RAW            & 0.592    & 0.607       & 0.594     & 0.576  & 0.585     & 0.592 \\ \cline{2-8} 
                      & GAN            & 0.498    & 0.767       & 0.494     & 0.229  & 0.313     & 0.498 \\ \cline{2-8} 
                      & aWOE           & 0.876    & 0.88        & 0.878     & 0.872  & 0.875     & 0.876 \\ \cline{2-8} 
\multirow{-4}{*}{LR}  & GAN-aWOE (avg) & 0.859    & 0.832       & 0.84      & 0.886  & 0.863     & 0.859 \\ \hline
                      & RAW            & 0.618    & 0.622       & 0.618     & 0.614  & 0.616     & 0.618 \\ \cline{2-8} 
                      & GAN            & 0.501    & 0.999       & 0.5       & 0.001  & 0.001     & 0.5   \\ \cline{2-8} 
                      & aWOE           & 0.867    & 0.87        & 0.869     & 0.863  & 0.866     & 0.867 \\ \cline{2-8} 
\multirow{-4}{*}{RF}  & GAN-aWOE (avg) & 0.749    & 0.901       & 0.858     & 0.598  & 0.701     & 0.749 \\ \hline
                      & RAW            & 0.548    & 0.554       & 0.547     & 0.541  & 0.544     & 0.548 \\ \cline{2-8} 
                      & GAN            & 0.497    & 0.222       & 0.497     & 0.773  & 0.605     & 0.498 \\ \cline{2-8} 
                      & aWOE           & 0.731    & 0.742       & 0.735     & 0.72   & 0.728     & 0.731 \\ \cline{2-8} 
\multirow{-4}{*}{DT}  & GAN-aWOE (avg) & 0.642    & 0.75        & 0.681     & 0.534  & 0.596     & 0.642 \\ \hline
                      & RAW            & 0.59     & 0.56        & 0.584     & 0.621  & 0.602     & 0.59  \\ \cline{2-8} 
                      & GAN            & 0.501    & 0.999       & 0.5       & 0.001  & 0.001     & 0.5   \\ \cline{2-8} 
                      & aWOE           & 0.866    & 0.87        & 0.868     & 0.861  & 0.865     & 0.866 \\ \cline{2-8} 
\multirow{-4}{*}{GB}  & GAN-aWOE (avg) & 0.738    & 0.818       & 0.79      & 0.658  & 0.71      & 0.738 \\ \hline
                      & RAW            & 0.537    & 0.578       & 0.539     & 0.495  & 0.516     & 0.537 \\ \cline{2-8} 
                      & GAN            & 0.514    & 0.778       & 0.528     & 0.249  & 0.339     & 0.513 \\ \cline{2-8} 
                      & aWOE           & 0.883    & 0.878       & 0.878     & 0.888  & 0.883     & 0.883 \\ \cline{2-8} 
\multirow{-4}{*}{FNN} & GAN-aWOE (avg) & 0.855    & 0.822       & 0.833     & 0.889  & 0.86      & 0.855 \\ \hline
                      & RAW            & 0.552    & 0.555       & 0.551     & 0.548  & 0.55      & 0.552 \\ \cline{2-8} 
                      & GAN            & 0.499    & 0.981       & 0.442     & 0.015  & 0.029     & 0.498 \\ \cline{2-8} 
                      & aWOE           & 0.714    & 0.718       & 0.712     & 0.709  & 0.711     & 0.714 \\ \cline{2-8} 
\multirow{-4}{*}{RNN} & GAN-aWOE (avg) & 0.738    & 0.761       & 0.755     & 0.715  & 0.728     & 0.738 \\ \hline
\end{tabular}
}
\end{center}
\end{table}

\begin{table}[]
\begin{center}
\caption{RAW, GANs, aWOE and GANs-aWOE based models on Dataset-2. For each performance metric, the best result is shown in \textbf{bold-face}.}
\label{table:GANs_WOE_methods_on_dataset-2}
{\scriptsize
\begin{tabular}{|l|l|l|l|l|l|l|l|}
\hline

Classifier            & MODEL          & Accuracy & Specificity & Precision & Recall & F-Measure & AUC   \\ \hline
                      & RAW            & 0.756    & 0.754       & 0.536     & 0.763  & 0.63      & 0.758 \\ \cline{2-8} 
                      & GAN            & 0.728    & 0.999       & 0.5       & 0.002  & 0.003     & 0.501 \\ \cline{2-8} 
                      & aWOE           & 0.814    & 0.903       & 0.68      & 0.568  & 0.619     & 0.736 \\ \cline{2-8} 
\multirow{-4}{*}{NB}  & GAN-aWOE (avg) & 0.832    & 0.907       & 0.718     & 0.632  & 0.671     & 0.769 \\ \hline
                      & RAW            & 0.772    & 0.94        & 0.667     & 0.321  & 0.433     & 0.63  \\ \cline{2-8} 
                      & GAN            & 0.724    & 0.989       & 0.32      & 0.014  & 0.027     & 0.501 \\ \cline{2-8} 
                      & aWOE           & 0.799    & 0.887       & 0.639     & 0.555  & 0.594     & 0.721 \\ \cline{2-8} 
\multirow{-4}{*}{KNN} & GAN-aWOE (avg) & 0.82     & 0.932       & 0.74      & 0.52   & 0.61      & 0.726 \\ \hline
                      & RAW            & 0.809    & 0.907       & 0.686     & 0.545  & 0.608     & 0.726 \\ \cline{2-8} 
                      & GAN            & 0.727    & 0.997       & 0.375     & 0.005  & 0.01      & 0.5   \\ \cline{2-8} 
                      & aWOE           & 0.821    & 0.918       & 0.709     & 0.552  & 0.621     & 0.735 \\ \cline{2-8} 
\multirow{-4}{*}{LR}  & GAN-aWOE (avg) & 0.828    & 0.88        & 0.682     & 0.691  & 0.686     & 0.785 \\ \hline
                      & RAW            & 0.805    & 0.927       & 0.708     & 0.477  & 0.57      & 0.702 \\ \cline{2-8} 
                      & GAN            & 0.272    & 0.001       & 0.271     & 0.998  & 0.427     & 0.499 \\ \cline{2-8} 
                      & aWOE           & 0.82     & 0.917       & 0.705     & 0.553  & 0.62      & 0.735 \\ \cline{2-8} 
\multirow{-4}{*}{RF}  & GAN-aWOE (avg) & 0.81     & 0.897       & 0.678     & 0.577  & 0.622     & 0.737 \\ \hline
                      & RAW            & 0.726    & 0.81        & 0.495     & 0.5    & 0.497     & 0.655 \\ \cline{2-8} 
                      & GAN            & 0.333    & 0.136       & 0.271     & 0.862  & 0.413     & 0.499 \\ \cline{2-8} 
                      & aWOE           & 0.762    & 0.849       & 0.554     & 0.521  & 0.537     & 0.685 \\ \cline{2-8} 
\multirow{-4}{*}{DT}  & GAN-aWOE (avg) & 0.771    & 0.877       & 0.606     & 0.486  & 0.522     & 0.682 \\ \hline
                      & RAW            & 0.796    & 0.929       & 0.696     & 0.439  & 0.538     & 0.684 \\ \cline{2-8} 
                      & GAN            & 0.272    & 0.001       & 0.271     & 0.998  & 0.427     & 0.499 \\ \cline{2-8} 
                      & aWOE           & 0.807    & 0.943       & 0.731     & 0.432  & 0.543     & 0.687 \\ \cline{2-8} 
\multirow{-4}{*}{GB}  & GAN-aWOE (avg) & 0.809    & 0.92        & 0.71      & 0.513  & 0.589     & 0.716 \\ \hline
                      & RAW            & 0.766    & 0.833       & 0.567     & 0.585  & 0.576     & 0.709 \\ \cline{2-8} 
                      & GAN            & 0.681    & 0.901       & 0.261     & 0.094  & 0.138     & 0.497 \\ \cline{2-8} 
                      & aWOE           & 0.82     & 0.904       & 0.688     & 0.587  & 0.634     & 0.746 \\ \cline{2-8} 
\multirow{-4}{*}{FNN} & GAN-aWOE (avg) & 0.823    & 0.87        & 0.668     & 0.696  & 0.681     & 0.783 \\ \hline
                      & RAW            & 0.767    & 0.966       & 0.718     & 0.235  & 0.354     & 0.6   \\ \cline{2-8} 
                      & GAN            & 0.728    & 0.999       & 0.5       & 0.002  & 0.003     & 0.501 \\ \cline{2-8} 
                      & aWOE           & 0.768    & 0.883       & 0.581     & 0.447  & 0.505     & 0.665 \\ \cline{2-8} 
\multirow{-4}{*}{RNN} & GAN-aWOE (avg) & 0.727    & 0.97        & 0.828     & 0.058  & 0.103     & 0.514 \\ \hline
\end{tabular}
}
\end{center}
\end{table}

\begin{table}[]
\begin{center}
\caption{RAW, GANs, aWOE and GANs-aWOE based models on Dataset-3. For each performance metric, the best result is shown in \textbf{bold-face}.}
\label{table:GANs_WOE_methods_on_dataset-3}
{\scriptsize
\begin{tabular}{|l|l|l|l|l|l|l|l|}
\hline

Classifier            & MODEL          & Accuracy & Specificity & Precision & Recall & F-Measure & AUC   \\ \hline
                      & RAW            & 0.856    & 0.922       & 0.477     & 0.444  & 0.46      & 0.683 \\ \cline{2-8} 
                      & GAN            & 0.922    & 0.982       & 0.83      & 0.546  & 0.659     & 0.764 \\ \cline{2-8} 
                      & aWOE           & 0.919    & 0.968       & 0.762     & 0.621  & 0.684     & 0.795 \\ \cline{2-8} 
\multirow{-4}{*}{NB}  & GAN-aWOE (avg) & 0.932    & 0.981       & 0.837     & 0.626  & 0.716     & 0.803 \\ \hline
                      & RAW            & 0.879    & 0.993       & 0.791     & 0.164  & 0.272     & 0.579 \\ \cline{2-8} 
                      & GAN            & 0.867    & 0.999       & 0.9       & 0.043  & 0.083     & 0.521 \\ \cline{2-8} 
                      & aWOE           & 0.926    & 0.986       & 0.867     & 0.56   & 0.68      & 0.773 \\ \cline{2-8} 
\multirow{-4}{*}{KNN} & GAN-aWOE (avg) & 0.868    & 0.999       & 0.892     & 0.045  & 0.085     & 0.522 \\ \hline
                      & RAW            & 0.594    & 0.853       & 0.522     & 0.227  & 0.316     & 0.54  \\ \cline{2-8} 
                      & GAN            & 0.908    & 0.999       & 0.986     & 0.343  & 0.509     & 0.671 \\ \cline{2-8} 
                      & aWOE           & 0.936    & 0.979       & 0.84      & 0.678  & 0.75      & 0.828 \\ \cline{2-8} 
\multirow{-4}{*}{LR}  & GAN-aWOE (avg) & 0.916    & 0.998       & 0.976     & 0.402  & 0.569     & 0.7   \\ \hline
                      & RAW            & 0.931    & 0.985       & 0.856     & 0.574  & 0.687     & 0.78  \\ \cline{2-8} 
                      & GAN            & 0.877    & 0.999       & 0.96      & 0.116  & 0.207     & 0.558 \\ \cline{2-8} 
                      & aWOE           & 0.931    & 0.984       & 0.862     & 0.608  & 0.713     & 0.796 \\ \cline{2-8} 
\multirow{-4}{*}{RF}  & GAN-aWOE (avg) & 0.866    & 0.999       & 0.791     & 0.031  & 0.059     & 0.515 \\ \hline
                      & RAW            & 0.906    & 0.938       & 0.646     & 0.705  & 0.674     & 0.822 \\ \cline{2-8} 
                      & GAN            & 0.881    & 0.966       & 0.621     & 0.348  & 0.446     & 0.657 \\ \cline{2-8} 
                      & aWOE           & 0.899    & 0.947       & 0.652     & 0.607  & 0.629     & 0.777 \\ \cline{2-8} 
\multirow{-4}{*}{DT}  & GAN-aWOE (avg) & 0.873    & 0.989       & 0.693     & 0.166  & 0.254     & 0.578 \\ \hline
                      & RAW            & 0.909    & 0.989       & 0.859     & 0.411  & 0.556     & 0.7   \\ \cline{2-8} 
                      & GAN            & 0.915    & 0.998       & 0.976     & 0.391  & 0.559     & 0.695 \\ \cline{2-8} 
                      & aWOE           & 0.934    & 0.979       & 0.835     & 0.661  & 0.738     & 0.82  \\ \cline{2-8} 
\multirow{-4}{*}{GB}  & GAN-aWOE (avg) & 0.871    & 0.999       & 0.941     & 0.072  & 0.133     & 0.536 \\ \hline
                      & RAW            & 0.849    & 0.978       & 0.237     & 0.043  & 0.073     & 0.511 \\ \cline{2-8} 
                      & GAN            & 0.905    & 0.999       & 0.985     & 0.314  & 0.476     & 0.657 \\ \cline{2-8} 
                      & aWOE           & 0.931    & 0.975       & 0.816     & 0.663  & 0.732     & 0.819 \\ \cline{2-8} 
\multirow{-4}{*}{FNN} & GAN-aWOE (avg) & 0.898    & 0.995       & 0.901     & 0.376  & 0.523     & 0.685 \\ \hline
                      & RAW            & 0.762    & 0.851       & 0.182     & 0.208  & 0.194     & 0.529 \\ \cline{2-8} 
                      & GAN            & 0.889    & 0.992       & 0.833     & 0.242  & 0.375     & 0.617 \\ \cline{2-8} 
                      & aWOE           & 0.899    & 0.991       & 0.859     & 0.345  & 0.492     & 0.668 \\ \cline{2-8} 
\multirow{-4}{*}{RNN} & GAN-aWOE (avg) & 0.896    & 0.973       & 0.721     & 0.411  & 0.519     & 0.692 \\ \hline
\end{tabular}
}
\end{center}
\end{table}

\end{document}